\documentclass{article} 
\usepackage{iclr2024_conference,times}

\usepackage{amsmath,amsfonts,bm}

\def\eqref#1{equation~\ref{#1}}

\def\1{\bm{1}}

\DeclareMathAlphabet{\mathsfit}{\encodingdefault}{\sfdefault}{m}{sl}
\SetMathAlphabet{\mathsfit}{bold}{\encodingdefault}{\sfdefault}{bx}{n}

\usepackage[utf8]{inputenc} 
\usepackage[T1]{fontenc}    
\usepackage{hyperref}       
\usepackage{url}            
\usepackage{booktabs}       
\usepackage{amsfonts}       
\usepackage{nicefrac}       
\usepackage{microtype}      
\usepackage{xcolor}         
 
\usepackage{microtype}
\usepackage{graphicx}
\usepackage{subfigure}
\usepackage{caption}
\usepackage{subcaption}
\usepackage{wrapfig}

\usepackage{amsfonts}
\usepackage{amsmath}
\usepackage{amsthm}
\usepackage{amssymb}
\usepackage{dsfont}
\usepackage{mathtools}
\usepackage{float}

\usepackage{algorithm}
\usepackage{algpseudocode}

\usepackage[capitalize,noabbrev]{cleveref}

\newtheorem{theorem}{Theorem}
\newtheorem{definition}{Definition}

\newtheorem{assumption}{Assumption}
\newtheorem{remark}{Remark}
\newtheorem{proposition}{Proposition}
\newtheorem{corollary}{Corollary}

\def\X{\mathcal{X}}
\newcommand{\norm}[1]{\left\lVert#1\right\rVert}

\def\frr{\mathrm{FRR}}
\def\trr{\mathrm{TRR}}
\def\far{\mathrm{FAR}}
\def\roc{\mathrm{ROC}}

\usepackage[textsize=tiny]{todonotes}

\title{Assessing Uncertainty in Similarity Scoring:\\ Performance \& Fairness in Face Recognition}

\author{Jean-Rémy Conti$^{1,2}$ \& Stéphan Clémençon$^{1}$\\
$^{1}$LTCI, Télécom Paris, Institut Polytechnique de Paris,
$^{2}$Idemia\\
\texttt{\{jean-remy.conti,stephan.clemencon\}@telecom-paris.fr}  \thanks{ Alternative correspondence: \texttt{jeanremy.conti@gmail.com}.}
}

\iclrfinalcopy 
\begin{document}

\maketitle

\begin{abstract}
The ROC curve is the major tool for assessing not only the performance but also the fairness properties of a similarity scoring function. In order to draw reliable conclusions based on empirical ROC analysis, accurately evaluating the uncertainty level related to statistical versions of the ROC curves of interest is absolutely necessary, especially for applications with considerable societal impact such as Face Recognition.  In this article, we prove asymptotic guarantees for empirical ROC curves of similarity functions as well as for by-product metrics useful to assess fairness. We also explain that, because the false acceptance/rejection rates are of the form of U-statistics in the case of similarity scoring, the naive bootstrap approach 
may 
jeopardize the assessment procedure. A dedicated recentering technique must be used instead. Beyond the theoretical analysis carried out, various experiments using real face image datasets provide strong empirical evidence of the practical relevance of the methods promoted here, when applied to several ROC-based measures such as popular fairness metrics.

\end{abstract}

\section{Introduction}\label{sec:introduction}

\begin{wrapfigure}[22]{r}{0.46\linewidth}
\centering
\vspace{-0.7cm}
\hspace*{0cm}
    \includegraphics[width=1.\linewidth]{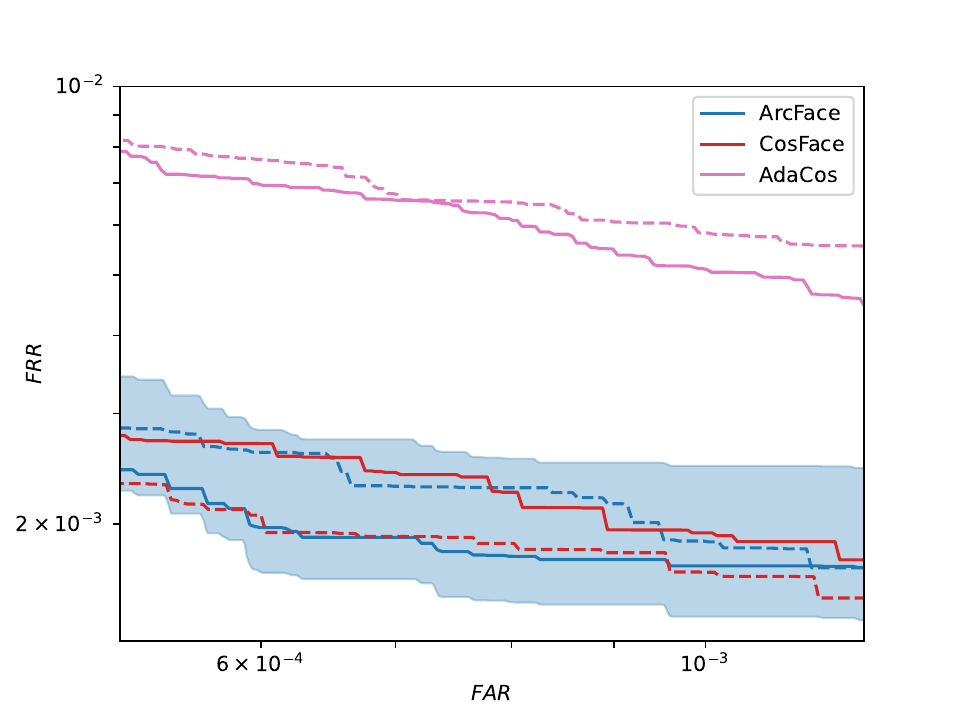}
    \caption{Empirical $\roc$ curves for three different models (ArcFace, CosFace, AdaCos) and for two distinct evaluation datasets (see~\ref{subsec:real_experiments}). The $\roc$ curves for the first dataset are depicted with solid lines while the $\roc$ curves for the second dataset are displayed with dashed lines. A confidence band for the $\roc$ computed with the ArcFace model on the first dataset is displayed in light blue. }
    \label{fig:motivation}
\end{wrapfigure}

The massive deployment of AI technologies brings with it a pressing demand for methodological tools to assess their trustworthiness. The reliability of AI systems concerns their estimated performance of course, but also their properties regarding fairness: ideally, the system should exhibit approximately the same performance, independently of the \textit{sensitive} group (determined by \textit{e.g.} gender, age group, race) to which it is applied. This is particularly true for Face Recognition (FR) systems, the running example through this article, now under scrutiny by the general public and regulatory organizations, refer to \textit{e.g.} \citet{Grother2019} or \citet{aclu_amazon_rekognition}.

The task of designing a FR system is usually formulated as a similarity scoring/learning problem, see \textit{e.g.} \cite{pmlr-v80-vogel18a}. Assuming that the system processes pixellated face images $X$ in $\mathbb{R}^d$, the goal is to build a (symmetric) scoring function $s:\mathbb{R}^d \times \mathbb{R}^d\to \mathbb{R}$ such that the larger the similarity score $s(x,x')$ related to a pair of images, the larger (hopefully) the probability that both images correspond to the same individual. In this case, one assigns a positive label to the pair and a negative label otherwise. The gold standard to measure the performance of such a similarity $s$ is the Receiver Operating Characteristic (ROC) curve \citep{GreSwe66}, namely the plot of the false rejection rate against the false acceptance rate, as the similarity scoring threshold varies. 

If, until now, the benchmark of FR systems has been essentially reduced to an ad-hoc evaluation of the performance metrics,  to the computation of empirical ROC curves based on FR evaluation datasets of reference (see \cite{Grother2019}), the quantification of the uncertainty inherent in the randomness of the evaluation datasets (referred to as \textit{aleatoric uncertainty} sometimes, see \cite{Eyke}) is essential to compare and appreciate fully the merits of such systems regarding accuracy and fairness with confidence.
The significance of the uncertainty quantification step is illustrated in Fig. \ref{fig:motivation}. The $\roc$ curves for three different FR models are computed on two distinct evaluation datasets (see \ref{subsec:real_experiments} for details on models and datasets). On the first dataset (solid lines), one would conclude that ArcFace is a better model than CosFace, as the former has a lower empirical $\frr$ than the latter, for any $\far$ value. However, one would draw the opposite conclusion, looking at the second dataset (dashed lines). Selecting models only depending on their empirical performance ($\roc$) has its flaws, as the uncertainty of those metrics is not taken into account. Note that the method for building confidence bands for the $\roc$, which we present in this paper, would have avoided such conclusions. Indeed, both models are indistinguishable in terms of performance, as their $\roc$ curves are contained within the band. On the contrary, the AdaCos model performs worse than ArcFace/CosFace on both datasets and its empirical $\roc$ curves are far from the confidence band. One could thus favor confidently ArcFace/CosFace over AdaCos regarding their performance.

In order to make meaningful comparisons, the (possibly high) uncertainty inherent in the statistical nature of the estimation must be taken into account. Indeed, this evaluation is crucial to judge whether the similarity scoring function candidates meet the performance/fairness requirements in a trustworthy manner. The main purpose of this paper is precisely to explain how to quantify the uncertainty/variability of similarity ROC curves, by means of a dedicated bootstrap methodology in particular, in a sound validity framework generalizing the one established by \cite{Bertail1} in the non pairwise setup.

{\bf Related works.}  To our knowledge, the uncertainty issue inherent to ROC curve estimation and fairness metrics estimation is poorly documented in the literature, particularly for similarity scoring problems such as FR. It has been studied at length for scoring functions, but in the non pairwise setup, using bootstrap methods in \citep{Bertail1}. The major difference between the analysis carried out therein and our framework lies in the fact that, in the similarity scoring context, false acceptance/rejection rates are not basic i.i.d. averages anymore but generalized $U$-statistics (refer to \cite{Lee1990} or \cite{ArconesGine92}). It has a significant impact on the methodology that can be used to quantify the uncertainty of empirical performance/fairness measures. As shall be seen in this paper, naively applying the bootstrap of \citet{Bertail1} to similarity scoring problems (\textit{e.g.} FR) strongly underestimates the $\roc$ curve, resulting in confidence bands for the $\roc$ curve which do not even contain the empirical $\roc$ curve most of the time. In \cite{pmlr-v80-vogel18a}, non asymptotic confidence bounds for the estimation error of empirical similarity $\roc$ curves have been established by means of linearization techniques tailored to $U$-statistics in a slightly different probabilistic framework (stipulating random labels). It is the purpose of the present paper to investigate how to accurately approximate the distribution of the estimation error by means of dedicated resampling techniques and build bootstrap confidence bands with satisfactory probability coverage. 

{\bf Contributions.} As will be explained in the subsequent analysis, a naive application of the bootstrap procedure yields a systematic underestimation of the similarity ROC curve. We provide \textit{(i)} a recentering technique to counteract this, while still being accurate asymptotically. Resulting from this bootstrap variant, \textit{(ii)} confidence bands for the $\roc$ curve and FR fairness metrics are shown to be consistent, in addition to achieve nominal coverage on synthetic data. The recentered bootstrap also allows to define \textit{(iii)} a scalar uncertainty measure for the $\roc$ and fairness metrics, which can be employed to compare the robustness of several FR fairness metrics. Finally, in addition to the statistical analysis presented, the relevance of the approach is supported by \textit{(iv)} illustrative numerical experiments, based on real data of face images, together with a discussion about the practical use of the information produced, in order to make more reliable decisions concerning accuracy and fairness. These results pave the way for a more valuable and trustworthy comparative analysis of the merits and drawbacks of FR systems.



{\bf Organization of the paper.} The main concepts at work in similarity scoring/learning and in FR are briefly recalled in Section \ref{sec:background}, together with the notions pertaining to ROC analysis used in this article to evaluate predictive performance and build fairness criteria. The consistency of empirical similarity ROC curves, is stated in section \ref{sec:uncertainty}. It is also explained therein how to bootstrap empirical similarity ROC curves in a valid manner, as well as by-product summary statistics reflecting the accuracy or fairness properties of the similarity scoring functions under study. Numerical experiments are presented and discussed in section \ref{subsec:real_experiments} for illustration purpose. 

\section{Background and Preliminaries}\label{sec:background}
We introduce here the main notations used throughout the article and briefly recall the standard similarity scoring/learning framework, involved in the design of FR systems in particular, and the key concepts pertaining to ROC analysis that are involved in the subsequent study. We next explain how to formulate fairness criteria based on similarity ROC curves in the FR context.
Here and throughout, the indicator function of any event $\mathcal{E}$ is denoted by $\mathbb{I}\{\mathcal{E}\}$, the Dirac mass at any point $x$ by $\delta_x$, and the pseudo-inverse of any cumulative distribution function (cdf) $\kappa(t)$ on $\mathbb{R}$ by $\kappa^{-1}(\alpha)=\inf\{t\in \mathbb{R}:\; \kappa(t)\geq \alpha  \}$.

\subsection{Similarity Scoring - Probabilistic and Statistical Framework}\label{subsec:framework}
The probabilistic framework considered here to formulate the similarity learning problem is the same as that of multi-class classification: $Y$ is a discrete random label defined on a probability space $(\Omega,\; \mathcal{A},\; \mathbb{P})$, valued in $\mathcal{Y}=\{1,\; \ldots,\; K\}$ with $K\geq 2$, and $X$ is a random vector defined on the same probability space and taking its values in a high dimensional space $\mathcal{X}\subset \mathbb{R}^d$ with $d \gg 1$. For all $k\in\mathcal{Y}$, we denote the supposedly continuous conditional distribution of $X$ given $Y=k$ by $F_k$, the probability that $Y$ equals $k$ by $p_k=\mathbb{P}\{Y=k\}$. Equipped with these notations, the joint distribution $P$ of the random pair $(X,Y)$ is fully characterized by $\{(F_k,p_k):\; k=1,\; \ldots,\; K\}$. 
In the running example considered through this paper, $X$ is an image depicting the face of an individual in a population of $K$ identities, the identity being indexed by $Y$.
In similarity learning, the objective pursued is to find a mapping $s:\mathcal{X}^2\to \mathbb{R}\cup\{+\infty\}$, called a \textit{similarity scoring function}, such that, given two independent pairs $(X,Y)$ and $(X',Y)$, the larger the similarity score $s(X,X')$, the more likely the same label should be shared (\textit{i.e.} one should observe the event $Y=Y'$) ideally. Before recalling performance/fairness metrics in similarity scoring, we explain the usual methodology at work in FR. 

{\bf Similarity scoring in Face Recognition.} In FR, one learns, from a dataset of face images with identity labels, an encoder function $f: \mathbb{R}^{h \times w \times c} \rightarrow \mathbb{R}^p$ that embeds the images in a way that brings same identities closer together in a certain sense. Each image is of size $(h,w)$, while $c$ corresponds to the color channel dimension. It is worth noticing that a pre-processing detection step (finding a face within an image) is required so that all face images have the same size $(h,w)$. For an image $x \in \mathbb{R}^{h \times w \times c}$, its latent representation $f(x) \in \mathbb{R}^p$ is referred to as the face embedding of $x$.
Since the advent of deep learning, the encoder $f$ is usually a deep Convolutional Neural Network (CNN) whose parameters are learned on a huge FR dataset, made of face images and identity labels. In brief, the training consists in taking all images $x_i^{(k)}$, labelled with identity $k$, computing their embeddings $f(x_i^{(k)})$ and adjusting the parameters of $f$ so that those embeddings are as close as possible (for a given similarity measure) and as far as possible from the embeddings of identity $l \neq k$. The usual similarity measure is the \textit{cosine similarity}, defined as 
\begin{equation}\label{eq:cos_sim}
s(x_i, x_j) := \frac{f(x_i)^\intercal f(x_j) }{  \norm{ f(x_i) } \cdot \norm{ f(x_j) } }
\end{equation}
for two images $x_i$ and $x_j$, where  $\norm{\cdot}$ stands for the usual Euclidean norm. In some early works~\citep{facenet}, the Euclidean metric $\norm{f(x_i) - f(x_j)}$ is also used. 


{\bf Multi-sample statistical setup.} In the subsequent analysis, the $p_k$'s are supposed to be given and are part and parcel of the performance criterion (a possible choice consists in giving the same weight to all identities, \textit{i.e.} $p_k=1/K$ for $k=1,\; \ldots,\; K$), whereas the $F_k$'s are unknown in practice. In order to evaluate empirically the properties of a (trained) similarity scoring function $s(x,x')$, it is assumed throughout the paper that $K$ i.i.d. samples are available: $$
X^{(k)}_{1},\; \ldots,\; X^{(k)}_{n_k}\overset{i.i.d.}{\sim} F_k,  \text{ where } n_k\geq 1 \text{ for } k=1,\; \ldots ,\; K .$$
In the FR context, we thus suppose that $n_k$ images are available for identity $k\in\{1,\; \ldots,\; K\}$,
the size of the pooled sample being denoted by $n=n_1+\ldots+n_K$. Those images belong to a test dataset, used to evaluate a trained FR model $f$ (or equivalently its associated similarity $s$ in Eq.~\ref{eq:cos_sim}). Based on these data, empirical versions of performance and fairness metrics can be computed. Their statistical accuracy will be next investigated from an asymptotic perspective, as the $n_k$'s simultaneously tend to infinity at the same rate as $n$: for all $k\in \mathcal{Y}$, there exists $\lambda_k>0$ such that
$n_k/n \rightarrow \lambda_k \text{ as } n\rightarrow +\infty$.

\subsection{ROC Analysis - Evaluation of Performance/Fairness in Similarity Scoring}\label{subsec:roc}


As formulated in \cite{pmlr-v80-vogel18a}, similarity learning can be seen as a specific \textit{bipartite ranking} problem, where the input space is of the form of a product space $\mathcal{X}\times \mathcal{X}$: given two independent observations $(X,Y)$ and $(X',Y')$ drawn from $\mathbb{P}$, the input r.v. is formed by the pair $(X,X')$, while $Z=2\mathbb{I}\{Y=Y'  \}-1$ is the binary label. The gold standard to evaluate bipartite ranking performance is $\roc$ analysis: a statistical learning view can be found in \textit{e.g.} \cite{CV09ieee}. In the similarity learning context, the $\roc$ curve of a similarity scoring function $s(x,x')$ is the plot of the False Rejection Rate (FRR) against the False Acceptance Rate (FAR) as the acceptance threshold varies, namely the mapping $\roc \colon \alpha \in (0,1)\mapsto \frr \circ \far^{-1}(\alpha)$,
where, for all $t\in \mathbb{R}$,
\begin{eqnarray*}
\far(t)&=&\mathbb{P}\{ s(X,X') > t  \mid Z=-1 \}
=\frac{\sum_{k< l} p_kp_l \mathbb{P}\{ s(X,X') > t  \mid Y=k,\; Y'=l \}}{\sum_{k< l}p_kp_l},\\
\frr(t)&=& \mathbb{P}\{ s(X,X')\leq t  \mid Z=+1 \}
= \frac{\sum_{k\in \mathcal{Y}}p_k^2\mathbb{P}\{ s(X,X')\leq t  \mid Y=Y'=k \}}{\sum_{k\in \mathcal{Y}}p_k^2}.
\end{eqnarray*}
A note on the definition of the pseudo-inverse for the $\far$ quantity is available in~\ref{app:inverse_far}.




\begin{remark}{\sc ($\roc$ conventions)}
In machine learning, the $\roc$ curve usually refers to the PP-plot $t\in \mathbb{R}\mapsto (\far(t),1-\frr(t))$, or equivalently $\alpha\in (0,1)\mapsto 1-\frr\circ\far^{-1}(\alpha)$. The FR community preferably plots $\far(t)$ on the $x$-axis and $\frr(t)$ on the $y$-axis. Both components correspond to error rates that should be minimized, one possibly more than the other depending on the use case. 
Of course, these two curves provide exactly the same information as there is a one-to-one correspondence between them. Note that we use the FR convention throughout the paper.
\end{remark}
In practice, special attention is paid to certain points of the $\roc$ curve.
The $\mathrm{FAR}$ level $\alpha\in (0,1)$ determines the operational point of the FR system and corresponds to the security risk one is ready to take. According to the FR use case, it is typically set to $10^{-i}$ with $i \in \{1, \ldots, 9\}$.

{\bf Fairness metrics.} In order to inspect the fairness properties of a FR system based on a similarity scoring function $s$, one generally looks at differentials in performance amongst several subgroups/segments of the population, a \textit{sensitive attribute} (\textit{e.g.} gender, race, age class, ...) making them distinguishable. For a given (discrete) sensitive attribute that can take $M > 1$ different values, in $\mathcal{A} = \{0, 1, \ldots, M-1 \}$ say, we enrich the probability space and now consider a random vector $(X,Y,A)$ where $A \in \mathcal{A}$ indicates the subgroup to which the individual indexed by $Y$ belongs to. For every fixed value $a\in \mathcal{A}$, we can further define the $\far$/$\frr$ related to subgroup $a$:
$\far_{a}(t)= \mathbb{P}\{ s(X,X') > t \ | \ Y \neq Y', \ A=A'=a\}$ and
$\frr_{a}(t)= \mathbb{P}\{ s(X,X') \leq t \ | \ Y = Y', \ A=A'=a\}$ for all $t\in \mathbb{R}$,
where by $(X',Y',A')$ is meant an independent copy of the random triplet $(X,Y,A)$. Ideally, a fair scoring function $s$ would exhibit nearly constant $\far_{a}(t)$ values when $a$ varies, for all $t  $ (and the same property for the $\frr_{a}(t)$ values). A FR fairness metric quantifies how much a model $s$ is far from this property. The FR fairness metrics considered in this paper are those used by the U.S. National Institute of Standards and Technology (NIST) in their FRVT report~\citep{frvt_part8}. They attempt to quantify the differentials in $(\mathrm{FAR}_a(t))_{a \in \mathcal{A}}$ and $(\mathrm{FRR}_a(t))_{a \in \mathcal{A}}$. Each fairness metric has two versions (one for the differentials in terms of $\mathrm{FAR}$, the other in terms of $\mathrm{FRR}$). 
A typical fairness metric is the max-min fairness below:
\[ \mathrm{FAR}_{\mathrm{min}}^\mathrm{max}(t) =  \frac{\max_{a \in \mathcal{A}} \mathrm{FAR}_a(t)}{\min_{a \in \mathcal{A}} \mathrm{FAR}_a(t)}, \quad \quad \mathrm{FRR}_{\mathrm{min}}^\mathrm{max}(t) =  \frac{\max_{a \in \mathcal{A}} \mathrm{FRR}_a(t)}{\min_{a \in \mathcal{A}} \mathrm{FRR}_a(t)}.\]
In practice, the threshold $t$ is set as for the $\roc$ curve, \textit{i.e.} it achieves a level $\far(t)=\alpha \in (0,1)$ for the global/total population, and not for some specific subgroup. Three other popular FR fairness metrics are the max-geomean metric, the log-geomean metric and the Gini coefficient. Their definition is postponed to \ref{app:fairness_metrics} for conciseness.

\section{Similarity Scoring Metrics - Assessing Uncertainty}\label{sec:uncertainty}

Motivated by the need to make trustworthy decisions taking into account the uncertainty inherent in the evaluation data, we now investigate the statistical accuracy of empirical counterparts of the $\roc$ curve of a given similarity scoring function $s(x,x')$ in the statistical multi-sample framework described in \ref{subsec:framework}. We next explain how to use the bootstrap methodology to estimate the related uncertainty level and build accurate confidence bands for the $\roc$ and fairness metrics. For notational simplicity, the results are stated and proved in the case where $p_k=1/K$ for all $k\in \mathcal{Y}$, extension to the general case being straightforward. 

\subsection{Statistical Inference - Consistency Result}\label{subsec:consistency}

An estimator of the $\roc$ curve is naturally obtained by replacing the quantities $\far$ and $\frr$ with their natural statistical counterparts in the definition of the $\roc$ curve. In the multi-sample statistical setup defined in \ref{subsec:framework}, the empirical versions of $\far(t)$ and $\frr(t)$ can be expressed as follows, using the symmetry property of similarity scoring functions: for all $t\in \mathbb{R}$,
\begin{eqnarray}\label{eq:emp_cdfH}
\widehat{\far}_{n}(t)&=&\frac{2}{K(K-1)}\sum_{k< l}\frac{1}{n_kn_l} \sum_{i=1}^{n_k}\sum_{j=1}^{n_l}\mathbb{I}\{ s(X^{(k)}_{i},X^{(l)}_{j}) > t  \},\\
\label{eq:emp_cdfG}
\widehat{\frr}_{n}(t)&=&\frac{1}{K}\sum_{k=1}^K\frac{2}{n_k(n_k-1)} \sum_{1\leq i<j\leq n_k}\mathbb{I}\{ s(X^{(k)}_{i},X^{(k)}_{j})\leq t  \}.
\end{eqnarray}
Notice that the terms involved in the two averages above are not independent, as each $X^{(k)}_{i}$ is involved in many terms of both averages, in contrast to the standard bipartite ranking framework \citep{Bertail1} where one deals with i.i.d. mean statistics. As detailed in \ref{app:u_stat_def}, these averages are actually generalized $U$-statistics, the simplest extensions of standard i.i.d. mean statistics. Properties and asymptotic theory of $U$-statistics can be found in \cite{Lee1990} while concentration properties are investigated in \cite{JMLRincompleteUstats}. The quantities (\ref{eq:emp_cdfH}) and (\ref{eq:emp_cdfG}) can then be used to compute the \textit{empirical similarity $\roc$ curve} based on the available evaluation datasets:
\begin{equation}\label{eq:ROC_emp}
\widehat{\roc}_n \colon \alpha \in (0,1)\mapsto \widehat{\frr}_{n} \circ (\widehat{\far}_{n})^{-1}(\alpha).
\end{equation}
Empirical versions of fairness metrics are naturally obtained in a similar \textit{plug-in} fashion (see \ref{app:empirical_fairness}).

The result stated below reveals the uniform consistency of the curve (\ref{eq:ROC_emp}) in the multi-sample asymptotic framework considered here.

\begin{proposition}\label{prop:limit}{\sc (Strong consistency)}
With probability one, we have:
\begin{equation}
\sup_{\alpha \in (0,1)} \{\widehat{\roc}_n(\alpha)- \roc(\alpha)\} \rightarrow 0, \text{ as } n\rightarrow +\infty.
\end{equation}
\end{proposition}
Refer to \ref{app:consistency_proof} for the technical proof. While the true $\roc$ curve is unknown, this result gives confidence in the quantity $\widehat{\roc}_n$ one computes based on data. A similar consistency result is stated in \cite{HT1}, when the negative and positive cdf's are estimated by basic i.i.d. averages, and proved by means of classic results for empirical processes. The case of empirical similarity $\roc$ curves cannot be handled in the same way because, as previously emphasized, (\ref{eq:emp_cdfH}) and (\ref{eq:emp_cdfG}) are generalized $U$-statistics and the terms involved in these averages exhibit a complex dependence structure. Linearization tricks (\textit{i.e.} Hoeffding decomposition), such as those used in \cite{pmlr-v80-vogel18a}, would be required to establish in addition the asymptotic Gaussianity of (a rescaled version of) the fluctuation process 
\begin{equation}\label{eq:fluctuat}
r_n(\alpha):=\sqrt{n} \{\widehat{\roc}_n(\alpha)- \roc(\alpha)\},\;\; \alpha \in (0,1).
\end{equation}
As underlined in \cite{Bertail1}, where the (much simpler) statistical framework considered is the same as in \cite{HT1},  the identification of the Gaussian limit of the process of Eq.~\ref{eq:fluctuat} is of very poor interest regarding the construction of (asymptotic) confidence bands for the similarity $\roc$ curve: beyond the computational difficulties inherent in simulating Brownian bridges, the presence of the unknown quantity $\roc(t)$ in the complex limit law makes its use impracticable to build confidence bands. Appropriate bootstrap techniques, whose asymptotic validity can be proved, should be preferably used instead.




\subsection{Bootstrapping the Performance/Fairness Metrics - Confidence Regions}\label{subsec:bootstrap}

Provided that representative datasets of the target populations are available, the empirical ROC curve~(\ref{eq:ROC_emp}) (and its related scalar summaries) of a similarity scoring function $s(x,x')$ is the main tool to assess performance and fairness in various applications such as FR. However, in order to make meaningful comparisons, the (possibly high) uncertainty inherent in the statistical nature of the estimation must be taken into account. Indeed, this evaluation is crucial to judge whether the similarity scoring function candidates meet the performance/fairness requirements in a trustworthy manner, as will be discussed on real examples in the next section. We now explain how to use a specific bootstrap resampling methodology to quantify the variability of the fluctuation process (\ref{eq:fluctuat}) in an asymptotically valid manner and why a naive bootstrap technique fails in the present situation.


{\bf Objective.} When computing $\widehat{\roc}_n(\alpha)$ to estimate the true $\roc$ curve, one makes the error 
\begin{equation}\label{eq:roc_error}
\hat{\epsilon}_n(\alpha) = \widehat{\roc}_n(\alpha) - \roc(\alpha),
\end{equation}
which is unknown, just like $\roc(\alpha)$. The variability of the random variable $\hat{\epsilon}_n(\alpha)$ fully characterizes the uncertainty of the empirical $\roc$ curve. The objective is to approximate $\hat{\epsilon}_n(\alpha)$ so that its variability can be estimated. This variability (\textit{i.e.} the uncertainty of $\widehat{\roc}_n(\alpha)$) will be used to build confidence bands around the empirical $\roc$ curve and to define a scalar uncertainty metric. In order to approximate $\hat{\epsilon}_n(\alpha)$, the bootstrap approach makes it possible to sample an estimate of $\hat{\epsilon}_n(\alpha)$. With many samples, one can retrieve the variability of the error $\hat{\epsilon}_n(\alpha)$.

 

\begin{figure}
\centering
\begin{minipage}{.46\textwidth}
  \centering
  \includegraphics[width=0.9\linewidth]{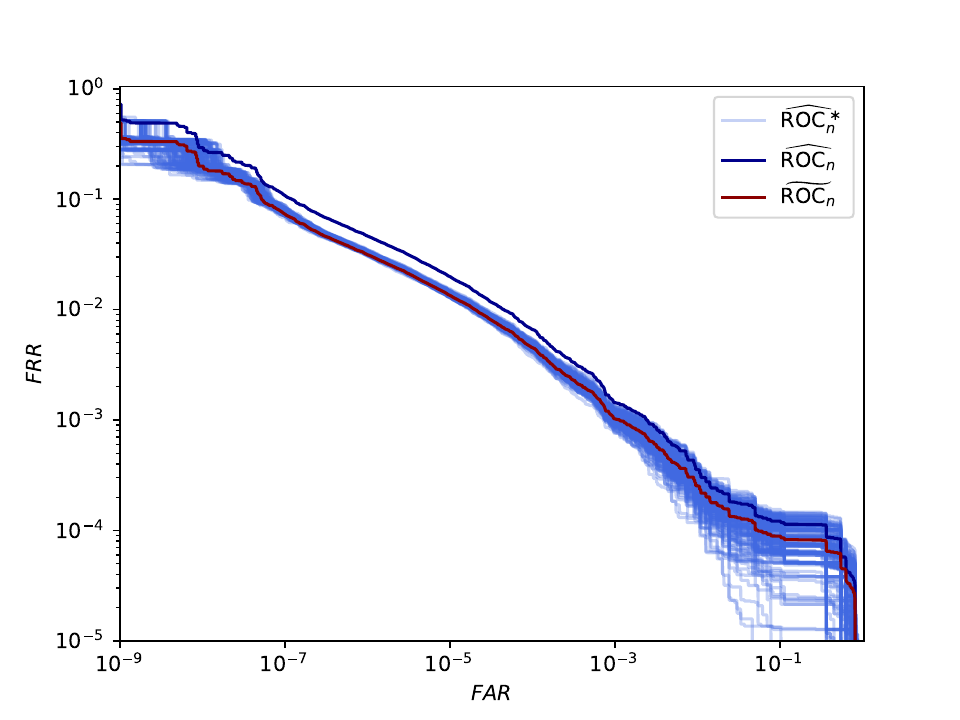}
  \captionof{figure}{Bootstrap versions of the ROC curve ($\widehat{\mathrm{ROC}}_n^*$ in light blue) and the empirical ROC curve ($\widehat{\mathrm{ROC}}_n$ in dark blue). The V-statistic counterpart $\widetilde{\mathrm{ROC}}_n$ is depicted in red.}
\label{fig:roc_bootstrap}
\end{minipage}%
\hspace{0.5cm}
\begin{minipage}{.46\textwidth}
  \centering
  \includegraphics[width=0.9\linewidth]{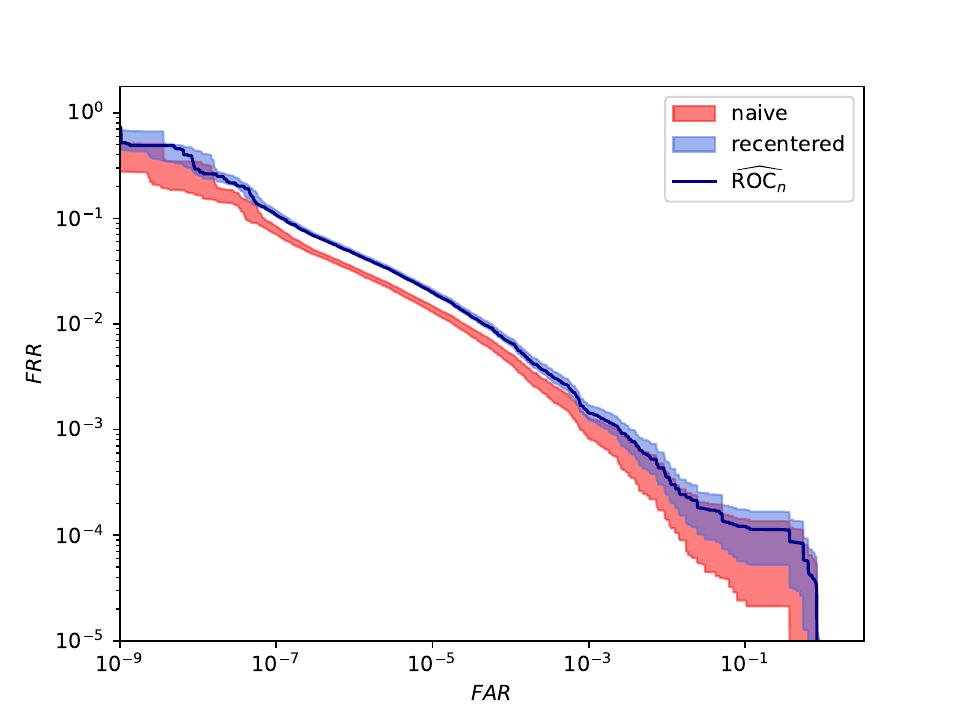}
  \captionof{figure}{Confidence bands at $95$\% confidence level for the empirical $\roc$ curve (dark blue), using two methods: the naive bootstrap (light red) and the recentered bootstrap (light blue).}
  \label{fig:naive_recentered_ci}
\end{minipage}
\end{figure}

{\bf Naive bootstrap.} The bootstrap paradigm, introduced by \cite{Efron} and developped at length in \citet{Bertail1}, suggests to recompute the empirical similarity $\roc$ curve (\ref{eq:ROC_emp}) from $K$ independent sequences of i.i.d. variables 
$X^{(k)*}_{1},\; \ldots,\; X^{(k)*}_{n_k}\overset{i.i.d.}{\sim} \hat{F}_k=\frac{1}{n_k}\sum_{1 \leq i\leq n_k}\delta_{X^{(k)}_i}$, conditioned upon the evaluation dataset $\mathcal{D}:=\{X^{(k)}_{1},\; \ldots,\; X^{(k)}_{n_k}:\; k=1,\; \ldots,\; K\}$. In other words, for each identity $k$ within the dataset, one simply randomly samples with replacement $n_k$ images $X^{(k)*}_{1},\; \ldots,\; X^{(k)*}_{n_k}$ among the $n_k$ face images $X^{(k)}_{1},\; \ldots,\; X^{(k)}_{n_k}$ available for identity $k$. The \textit{bootstrap sample}, which should be viewed as a sort of replicate of the evaluation dataset, is obtained by concatenating all the images thus sampled. Notice that the bootstrap sample may contain the same images several times due to the use of the sampling with replacement scheme, in contrast to the original dataset $\mathcal{D}$. 
In practice, the resampling scheme is replicated $B\geq 1$ times in order to compute a Monte-Carlo approximation of the distribution of the \textit{bootstrap $\roc$}, \textit{i.e.} the curve $\alpha \mapsto \widehat{\roc}_n^*(\alpha):=\widehat{\frr}^*_{n} \circ (\widehat{\far}^*_{n})^{-1}(\alpha) $, with
\begin{eqnarray}\label{eq:boot_far}
\widehat{\far}^*_{n}(t)&=&\frac{2}{K(K-1)}\sum_{k< l}\frac{1}{n_kn_l} \sum_{i=1}^{n_k}\sum_{j=1}^{n_l}\mathbb{I}\{ s(X^{(k)*}_{i},X^{(l)*}_{j}) > t  \},\\
\label{eq:boot_frr}
\widehat{\frr}^*_{n}(t)&=&\frac{1}{K}\sum_{k=1}^K\frac{2}{n_k(n_k-1)} \sum_{1\leq i<j\leq n_k}\mathbb{I}\{ s(X^{(k)*}_{i},X^{(k)*}_{j})\leq t  \}.
\end{eqnarray}
The curve $\widehat{\roc}_n^*$ is nothing but the empirical ROC curve computed with a bootstrap sample, instead of the original dataset. It turns out that, conditionally to the dataset $\mathcal{D}$, the quantity $\hat{\epsilon}_n^{(1)}(\alpha)~=~\widehat{\roc}_n^*(\alpha)-\widehat{\roc}_n(\alpha)$ approximates $\hat{\epsilon}_n(\alpha)$'s law. The approximation procedure is asymptotically valid, as proved in \ref{app:validity_bootstrap}, so that it satisfies our objective.
In practice, one would use $B$ bootstrap samples, get $B$ realizations of $\widehat{\roc}_n^*(\alpha)$, and thus $B$ realizations of $\hat{\epsilon}_n^{(1)}(\alpha)$ which allow to compute the variability of $\hat{\epsilon}_n^{(1)}(\alpha)$, in order to estimate the variability of $\hat{\epsilon}_n(\alpha)$, \textit{i.e.} the uncertainty of the $\roc$ curve. The \textit{bootstrap percentile} method enables to build accurate confidence regions for the quantity $\widehat{\roc}_n$ of interest. Indeed, for a fixed $\alpha \in (0,1)$, a confidence interval, at level $1-\alpha_{CI} \in [0,1]$, around $\widehat{\roc}_n(\alpha)$ can be obtained by considering the $\frac{\alpha_{CI}}{2}$-th and the $\frac{1-\alpha_{CI}}{2}$-th quantiles of $B$ realizations of $\widehat{\roc}_n(\alpha) + \hat{\epsilon}_n^{(1)}(\alpha)$.
It is worth noticing that, in spite of its asymptotic validity, the method can be seriously compromised in the non asymptotic regime when the distribution of $\widehat{\roc}_n^*(\alpha)$ is not centered at $\widehat{\roc}_n(\alpha)$. This is typically the case in the present situation, as depicted in Fig. \ref{fig:roc_bootstrap}: the $B=200$ realizations of $\widehat{\roc}_n^*(\alpha)$ (light blue) are not at all centered around the empirical $\roc$ curve (dark blue), the confidence band formed by the \textit{naive bootstrap}, at level $1-\alpha_{CI}=95\%$, being displayed in Fig. \ref{fig:naive_recentered_ci} (light red). Note that the naive bootstrap of \cite{Bertail1} yields strongly inaccurate confidence bands (in the non asymptotic regime) as the bands do not even contain the empirical $\roc$ curve most of the time. For both Figures, the pretrained encoder $f$ defining the similarity function $s$ (\textit{cf} Eq.~\ref{eq:cos_sim}) is ArcFace, while MORPH is the evaluation dataset (see \ref{subsec:real_experiments} for details on models and datasets). We now explain why the distribution of $\widehat{\roc}_n^*(\alpha)$ is not centered around $\widehat{\roc}_n(\alpha)$ \textit{i.e.} why the naive bootstrap of \cite{Bertail1} fails.


{\bf $V$-statistic version of the $\roc$ and recentering.}
Firstly, one can easily check that $\mathbb{E}^*[\widehat{\far}^*_{n}(t)\mid~\mathcal{D}]~=~\widehat{\far}_{n}(t)$, where by $\mathbb{E}^*[\cdot \mid \mathcal{D}]$ and $\mathbb{P}^*\{\cdot \mid \mathcal{D}\}$ 
are meant the conditional expectation 
and probability 
given the dataset $\mathcal{D}$ used to compute the empirical criterion $\widehat{\roc}_n$. Whereas the quantity $\widehat{\far}^*_{n}(t)$ is well centered around the empirical $\widehat{\far}_{n}(t)$ given $\mathcal{D}$,
this is not the case for the $\frr$ metric in general. Indeed, we have:
\begin{equation}\label{eq:Vstat}
\mathbb{E}^*\left[\widehat{\frr}^*_{n}(t)\mid \mathcal{D}\right]=\frac{1}{K}\sum_{k=1}^K\frac{1}{n^2_k} \sum_{1\leq i,j\leq n_k}\mathbb{I}\{ s(X^{(k)}_{i},X^{(k)}_{j})\leq t  \}:=\widetilde{\frr}_{n}(t).
\end{equation}
The quantity above is an average of $K$ independent $V$-statistics, that may slightly differ from $\widehat{\frr}_{n}(t)$, the difference being of order $O(1/n)$. This is due to the presence of the \textit{diagonal terms} $\mathbb{I}\{ s(X^{(k)}_{i},X^{(k)}_{i})\leq t  \}$, which results from the fact that
perfect similarities (\textit{i.e.} similarities equal to $1$ in the cosine similarity (\ref{eq:cos_sim})) can be observed in the bootstrap samples with non zero probability, while this cannot occur when computing $\widehat{\frr}_{n}(t)$. Hence, the empirical $\frr$ tends to be underestimated by its naive boostrap version in general. This is reflected in the bootstrap $\roc$ curves $\widehat{\roc}^*_{n}$ being centered around their V-statistic version $\widetilde{\roc}_n(\alpha):=
\widetilde{\frr}_{n} \circ 
(\widehat{\far}_{n})^{-1}(\alpha)$, and not around the empirical $\roc$ curve $\widehat{\roc}_n$, as depicted by Fig. \ref{fig:roc_bootstrap}. Notice incidentally that this phenomenon is specific to similarity scoring, because of the pairwise nature of the statistic (\ref{eq:emp_cdfG}), and does not occur in the classic bipartite ranking framework \citep{Bertail1}. In fact, the V-statistic version of the $\roc$ curve is of great interest since the recentered error $\hat{\epsilon}_n^{(2)}(\alpha) \colon=\widehat{\roc}_n^*(\alpha)-\widetilde{\roc}_n(\alpha)$ also approximates $\hat{\epsilon}_n(\alpha)$'s law, as proved in \ref{app:validity_bootstrap}. In the same way than for $\hat{\epsilon}_n^{(1)}(\alpha)$, we are able to build confidence intervals, at confidence level $1-\alpha_{CI} \in [0,1]$, for the quantity $\widehat{\roc}_n(\alpha)$. Considering $B$ bootstrap samples, one would compute $l_{\alpha_{CI}}^{(n,B)}(\alpha)$ (resp. $u_{\alpha_{CI}}^{(n,B)}(\alpha)$) the $\frac{\alpha_{CI}}{2}$-th (resp. the $\frac{1-\alpha_{CI}}{2}$-th) quantile of the $B$ realizations of $\widehat{\roc}_n(\alpha) + \hat{\epsilon}_n^{(2)}(\alpha)$. $l_{\alpha_{CI}}^{(n,B)}(\alpha)$ and $u_{\alpha_{CI}}^{(n,B)}(\alpha)$ are respectively the lower and upper bounds of the confidence interval, obtained with a variant of the naive bootstrap which we call $\textit{recentered bootstrap}$. The difference with $\hat{\epsilon}_n^{(1)}(\alpha)$ is that, using Eq. \ref{eq:Vstat}, one finds that $\mathbb{E}^*\left[\widehat{\roc}_{n}(\alpha) + \hat{\epsilon}_n^{(2)}(\alpha) \mid \mathcal{D}\right] = \widehat{\roc}_{n}(\alpha)$, \textit{i.e.} the confidence intervals are well centered around the empirical $\roc$ curve, as depicted in Fig.~\ref{fig:naive_recentered_ci}. 

The theoretical considerations within this section can be summarized by the result below. It states that the confidence interval, at confidence level $1-\alpha_{CI}$, for the $\roc$ curve, using the recentered bootstrap, has a probability of containing the true $\roc$ curve which is truly equal to $1-\alpha_{CI}$ when $n$ and $B$ tend to $+\infty$. Refer to \ref{app:consistency_ci} for the technical proof. The result holds when applying the recentered bootstrap to the considered fairness metrics (see~\ref{app:fairness_bootstrap}).

\begin{theorem}\label{thm:asym_coverage}
Let $\alpha\in (0,1)$ and $\alpha_{CI}\in (0,1)$. Under the (mild) assumptions in \ref{app:assumptions}, we have:
\[
\mathbb{P}\{ l_{\alpha_{CI}}^{(n,B)}(\alpha) \leq \roc(\alpha) \leq u_{\alpha_{CI}}^{(n,B)}(\alpha)\} \rightarrow 1-\alpha_{CI},
\]
as $n$ and $B$ both tend to $+\infty$.
\end{theorem}
This is an asymptotic result. In the non asymptotic regime, \textit{i.e.} with a finite evaluation dataset, an interesting question is to find what happens to this probability of containing the true $\roc$ curve. For that purpose, it is common practice to use synthetic datasets allowing for an approximation of the true $\roc$ curve. On each dataset, one can compute confidence intervals and check whether the frequency, over all datasets, of containing the true $\roc$ curve is truly equal to $1-\alpha_{CI}$. In \ref{app:coverage_experiment}, we generate $200$ synthetic datasets and conclude that it is very close to $1-\alpha_{CI}$, underlining the soundness of the recentered bootstrap, while it is not the case for the naive bootstrap of \citet{Bertail1}.

{\bf Uncertainty metric.} To quantify the uncertainty about the $\roc$ curve, one might be interested in a scalar quantity which summarizes its uncertainty. For instance, $\sqrt{\mathrm{Var}[\hat{\epsilon}_n(\alpha)]}$ seems appropriate for such a quantification. However, in order to compare this scalar uncertainty at several $\far$ levels $\alpha$, it seems reasonable to consider a \textit{relative} quantity such as $\sqrt{\mathrm{Var}[\hat{\epsilon}_n(\alpha)]} / \widehat{\roc}_n(\alpha)$. To estimate this quantity, we use the recentered bootstrap and define the \textit{normalized uncertainty} of the $\roc$ curve as:
\begin{equation}\label{eq:normalized_uncertainty}
U[\widehat{\roc}_n(\alpha)]  = \frac{\sqrt{\mathrm{Var}[\hat{\epsilon}_n^{(2)}(\alpha)\mid \mathcal{D}]}}{\widehat{\roc}_n(\alpha)}, 
\end{equation}
as $\hat{\epsilon}_n^{(2)}(\alpha) =\widehat{\roc}_n^*(\alpha)-\widetilde{\roc}_n(\alpha)$ approximates $\hat{\epsilon}_n(\alpha)$'s law. In practice, one would use $B$ bootstrap samples, get $B$ realizations of $\widehat{\roc}_n^*(\alpha)$, thus of $\hat{\epsilon}_n^{(2)}(\alpha)$. From those data, one would compute their standard deviation, normalized by the empirical $\roc$ curve. The definition of the normalized uncertainty is naturally extended to fairness metrics (see~\ref{app:fairness_bootstrap}).
 
 The pseudo-codes for the naive/recentered bootstrap methods, the computation of confidence intervals, as well as of the normalized uncertainty for the ROC curve and fairness metrics are available in~\ref{app:code}.

\section{Numerical Experiments - Applications}\label{subsec:real_experiments}


{\bf Models and datasets.}
We take as encoder $f$ several pre-trained models\footnote{\label{footnote:faceX_zoo}\url{https://github.com/JDAI-CV/FaceX-Zoo/blob/main/training_mode/README.md}.} (AdaCos of \cite{adacos}, ArcFace of \cite{arcface}, CosFace of \cite{cosface}, CurricularFace of \cite{curricularface}) whose backbone is a MobileFaceNet~\citep{mobilefacenets}, trained on the MS-Celeb-1M-v1c-r dataset\footnote{See footnote \ref{footnote:faceX_zoo}.}. This dataset is a cleaned version of the MS-Celeb1M dataset \citep{ms-celeb-1m} and it contains $3.28$M images of $73$k identities. We choose the MORPH dataset \citep{morph} as evaluation dataset. It is composed of $55$k face images from $13$k distinct identities. This dataset is widely used for fairness evaluation since it is provided with ground-truth age and gender labels (the available labels for the latter are female and male). All images are pre-processed by the Retina-Face detector \citep{retinaface_detector} and are of size $112\times112$ pixels. Unless specified, all experiments use $B=200$ bootstrap samples.

To highlight the significance of the tackled problem in this paper, we show in Fig.~\ref{fig:motivation} the empirical $\roc$ curves of three models (ArcFace, CosFace and AdaCos), computed on two distinct datasets. Those datasets are obtained by splitting MORPH in two parts, with the same number of images, each identity being present in both splits (see~\ref{app:morph_split}). A confidence band at $95\%$ confidence level, computed for ArcFace on one split of data, suggests that ArcFace and CosFace are indistinguishable in terms of performance, for the $\far$ levels displayed on the $x$-axis. This insight is interesting as there is no model between ArcFace and CosFace that performs better than the other on both datasets.  

\begin{figure}
\centering
\begin{minipage}{.46\textwidth}
  \centering
  \includegraphics[width=0.9\linewidth]{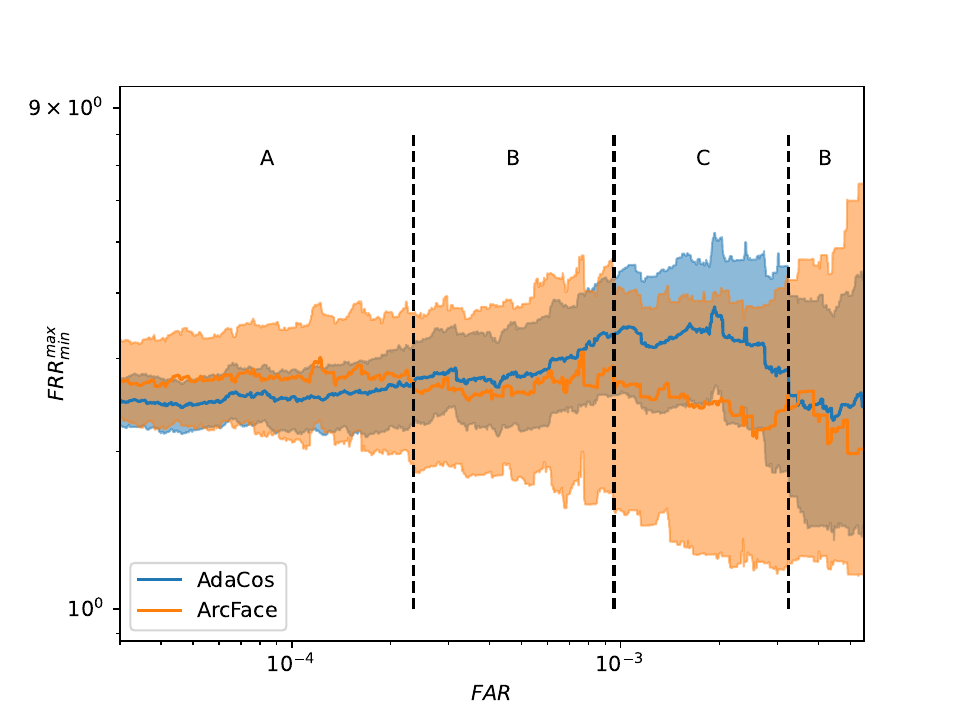}
  \captionof{figure}{Confidence bands at $95$\% confidence level for the $\mathrm{FRR}_{\mathrm{min}}^\mathrm{max}$ fairness metric, for two models (ArcFace, AdaCos). The empirical fairness metrics are depicted as solid lines.}
  \label{fig:method_decision_fairness}
\end{minipage}%
\hspace{0.5cm}
\begin{minipage}{.46\textwidth}
  \centering
  \includegraphics[width=0.9\linewidth]{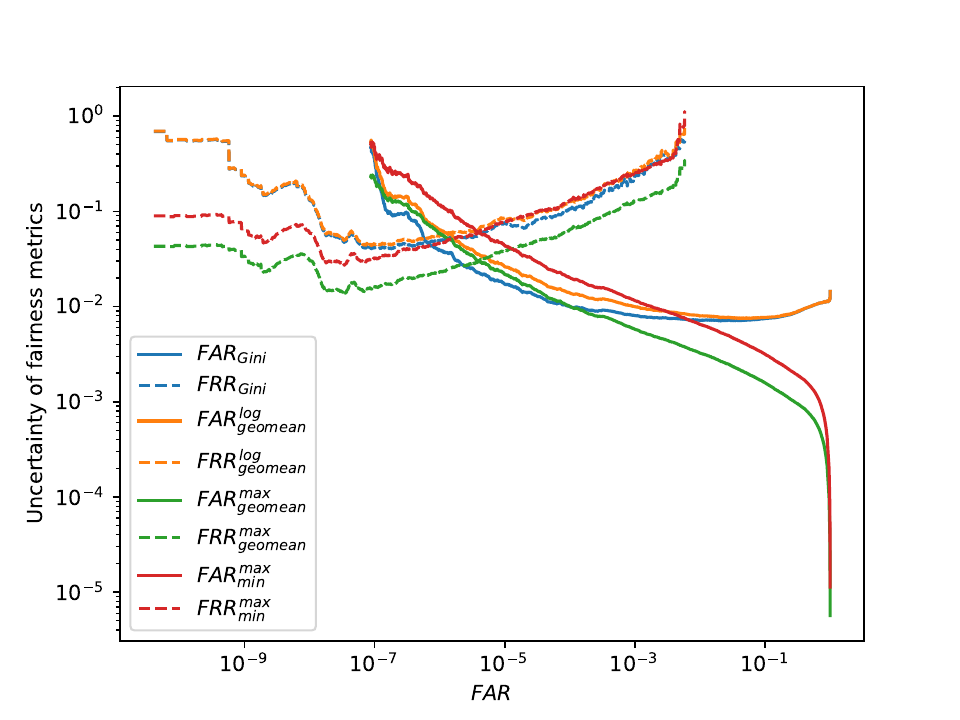}
  \captionof{figure}{Normalized uncertainty of several fairness metrics ($\far$ fairness in solid lines, $\frr$ fairness in dashed lines). The gender label is chosen as the sensitive attribute.}
  \label{fig:std_fairness}
\end{minipage}
\end{figure}

Then, we investigate the uncertainty related to the fairness metric $\frr_{\mathrm{min}}^{\mathrm{max}}$. The gender label is used here as the sensitive attribute. We display in Figure~\ref{fig:method_decision_fairness} the confidence bands at $95\%$ confidence level for the $\mathrm{FRR}_{\mathrm{min}}^\mathrm{max}$ fairness metric (see \ref{app:fig4_analogues} for other fairness metrics), for two models (AdaCos and ArcFace). Three zones (A, B, C) are delimited by dashed lines. For the zone A (resp. C), the empirical fairness is better for AdaCos (resp. ArcFace), while the upper-bound of the confidence band is lower for AdaCos (resp. ArcFace). One would conclude that, for each zone, one model is better than the other in terms of $\frr$ fairness (AdaCos for zone A, ArcFace for zone C). The case of zone B is more complex. Only using the empirical fairness metrics, one would choose ArcFace as the fair model. However, the uncertainty for ArcFace is high, and one may choose AdaCos for its robustness, especially in the case where there would be a strict fairness constraint to deploy the technology (\textit{e.g.} a legislation requiring $\mathrm{FRR}_{\mathrm{min}}^\mathrm{max} \leq 4$ at $\far=6\times 10^{-4}$ for any evaluation dataset).

Finally, we compute the normalized uncertainty of Eq.~\ref{eq:normalized_uncertainty} for all fairness metrics. As illustrated in Figure~\ref{fig:std_fairness}, the max-geomean metric displays (almost always) the lowest uncertainty, both in terms of $\mathrm{FAR}$ and $\mathrm{FRR}$, which makes it particularly suitable for fairness evaluation. This finding is supported by similar experiments in \ref{app:normalized_uncertainty_fairness_expes}, where the trained model, the evaluation dataset and the used sensitive attribute change. In particular, we employ a ArcFace model with a ResNet backbone, evaluated on RFW \citep{RFW}. In addition to be more robust than other fairness metrics, the max-geomean metric has the significant advantage to be interpretable. 



\section{Conclusion}\label{sec:conclusion}
In this paper, we consider the problem of assessing the uncertainty inherent in estimating $\roc$ curves, using evaluation datasets, in the context of similarity scoring. Quantifying this uncertainty is of great interest since the ROC curve is the gold standard to evaluate performance and fairness in various applications, such as Face Recognition. We show the consistency of empirical similarity ROC curves and propose a variant of the bootstrap approach to build confidence bands around performance/fairness metrics, in order to quantify their variability. The procedure is proved to be asymptotically valid and its relevance is illustrated through applications in Face Recognition. Finally, some popular Face Recognition fairness metrics are compared in terms of their uncertainty, revealing that the max-geomean metric is the more robust to assess fairness. While the gold standard by which fairness will be evaluated in the future is not fixed yet, we believe that it should definitely incorporate uncertainty measures, since it could lead to wrong conclusions otherwise. The bootstrap approach is simple, fast and could greatly improve the reliability of accuracy and fairness metrics, especially within the Face Recognition community.  

{\bf Reproducibility.} Pseudo-codes for the experiments of the paper are available in the Supplementary Material \ref{app:code}. The open-source pre-trained models used for the experiments are available with download links.



\subsubsection*{Acknowledgments}This research was partially supported by the French National Research Agency (ANR), under grant ANR-20-CE23-0028 (LIMPID project).

\bibliographystyle{iclr2024_conference}
\bibliography{biblio}

\newpage

\newpage
\appendix

\section{Further Remarks}

\subsection{Fairness Metrics}\label{app:fairness_metrics}

In order to inspect the fairness properties of a FR system based on a similarity scoring function $s$, one generally looks at differentials in performance amongst several subgroups/segments of the population. Such subgroups are distinguishable by a \textit{sensitive attribute} (\textit{e.g.} gender, race, age class, ...). For a given (discrete) sensitive attribute that can take $M > 1$ different values, in $\mathcal{A} = \{0, 1, \ldots, M-1 \}$ say, we enrich the probability space and now consider a random vector $(X,Y,A)$ where $A \in \mathcal{A}$ indicates the subgroup to which the individual indexed by $Y$ belongs to. For every fixed value $a\in \mathcal{A}$, we can further define the $\far$/$\frr$ related to subgroup $a$:
$\far_{a}(t)= \mathbb{P}\{ s(X,X') > t \ | \ Y \neq Y', \ A=A'=a\}$ and
$\frr_{a}(t)= \mathbb{P}\{ s(X,X') \leq t \ | \ Y = Y', \ A=A'=a\}$ for all $t\in \mathbb{R}$,
where by $(X',Y',A')$ is meant an independent copy of the random triplet $(X,Y,A)$. Ideally, a fair model/function $s$ would exhibit nearly constant $\far_{a}(t)$ values when $a$ varies, for all $t  $ (and the same property for the $\frr_{a}(t)$ values). A FR fairness metric quantifies how much a model $s$ is far from this property.

In the following, we list several popular FR fairness metrics. All of them are used by the U.S. National Institute of Standards and Technology (NIST) in their FRVT report~\citep{frvt_part8}. Those fairness metrics attempt to quantify the differentials in $(\mathrm{FAR}_a(t))_{a \in \mathcal{A}}$ and $(\mathrm{FRR}_a(t))_{a \in \mathcal{A}}$. Each fairness metric has two versions : one for the differentials in terms of $\mathrm{FAR}$, the other for the differentials in terms of $\mathrm{FRR}$. 

\paragraph{Max-min ratio.} This metric has also been introduced by \citet{JR_vmf}. Its advantage is to be very interpretable but it is sensitive to low values in the denominator.
\begin{align*}    
\mathrm{FAR}_{\mathrm{min}}^\mathrm{max}(t) &=  \frac{\max_{a \in \mathcal{A}} \mathrm{FAR}_a(t)}{\min_{a \in \mathcal{A}} \mathrm{FAR}_a(t)},\\
\mathrm{FRR}_{\mathrm{min}}^\mathrm{max}(t) &=  \frac{\max_{a \in \mathcal{A}} \mathrm{FRR}_a(t)}{\min_{a \in \mathcal{A}} \mathrm{FRR}_a(t)}.
\end{align*}

\paragraph{Max-geomean ratio.} This metric replaces the previous minimum by the geometric mean $\mathrm{FAR}^\dag(t)$ of the values $(\mathrm{FAR}_a(t))_{a \in \mathcal{A}}$, in order to be less sensitive to low values in the denominator.
\begin{align*}    
\mathrm{FAR}_{\mathrm{geomean}}^\mathrm{max}(t) &=  \frac{\max_{a \in \mathcal{A}} \mathrm{FAR}_a(t)}{\mathrm{FAR}^\dag(t)}, \\
\mathrm{FRR}_{\mathrm{geomean}}^\mathrm{max}(t) &=  \frac{\max_{a \in \mathcal{A}} \mathrm{FRR}_a(t)}{\mathrm{FRR}^\dag(t)}.
\end{align*}

\paragraph{Log-geomean sum.} It is a sum of normalized logarithms.
\begin{align*}    
 \mathrm{FAR}_{\mathrm{geomean}}^\mathrm{log}(t) &= \sum_{a \in \mathcal{A}} \left|\log_{10}       \frac{\mathrm{FAR}_a(t)}{\mathrm{FAR}^\dag(t)}\right|, \\
 \mathrm{FRR}_{\mathrm{geomean}}^\mathrm{log}(t) &= \sum_{a \in \mathcal{A}} \left|\log_{10}       \frac{\mathrm{FRR}_a(t)}{\mathrm{FRR}^\dag(t)}\right|.
 \end{align*}

\paragraph{Gini coefficient.} The Gini coefficient is a measure of inequality in a population. It  ranges from a minimum value of zero, when all individuals are equal, to a theoretical maximum of one in an infinite population in which every individual except one has a size of zero. In this case, the mean $\mathrm{FAR}^\diamond(t)$ that is used is the arithmetic mean of the values $(\mathrm{FAR}_a(t))_{a \in \mathcal{A}}$.
\begin{align*}    
\mathrm{FAR}_{\mathrm{Gini}}(t) &= \frac{\left|\mathcal{A}\right|}{\left|\mathcal{A}\right|-1} \displaystyle \frac{\sum_{a \in \mathcal{A}} \sum_{b \in \mathcal{A}} \left|\mathrm{FAR}_a(t) - \mathrm{FAR}_b(t)\right| }{2 \left|\mathcal{A}\right|^2 \mathrm{FAR}^\diamond(t)}, \\
 \mathrm{FRR}_{\mathrm{Gini}}(t) &= \frac{\left|\mathcal{A}\right|}{\left|\mathcal{A}\right|-1} \displaystyle \frac{\sum_{a \in \mathcal{A}} \sum_{b \in \mathcal{A}} \left|\mathrm{FRR}_a(t) - \mathrm{FRR}_b(t)\right| }{2 \left|\mathcal{A}\right|^2 \mathrm{FRR}^\diamond(t)}.
 \end{align*}

Note that it is common to plot those fairness metrics not as functions of the threshold $t$ but as functions of the level of $\far$ associated to the threshold (as for the $\roc$ curve), meaning that the threshold~$t$ is set so that it achieves a $\mathrm{FAR}(t) = \alpha \in (0,1)$ (for the global/total population, and not for some specific subgroup). In this sense, one would replace the threshold $t$ in each fairness metric by $\far^{-1}(\alpha)$. For instance, the strict definition of the max-min fairness metric is, for any $\alpha \in (0,1)$:
\begin{align*}    
\mathrm{FAR}_{\mathrm{min}}^\mathrm{max}(\alpha) &=  \frac{\max_{a \in \mathcal{A}} \mathrm{FAR}_a \circ \far^{-1}(\alpha)}{\min_{a \in \mathcal{A}} \mathrm{FAR}_a \circ \far^{-1}(\alpha)},\\
\mathrm{FRR}_{\mathrm{min}}^\mathrm{max}(\alpha) &=  \frac{\max_{a \in \mathcal{A}} \mathrm{FRR}_a \circ \far^{-1}(\alpha)}{\min_{a \in \mathcal{A}} \mathrm{FRR}_a \circ \far^{-1}(\alpha)}.
\end{align*}

For the sake of clarity, we listed the fairness metrics without the pseudo-inverse $\far^{-1}$. However, the theoretical results within this paper take into account those strict definitions.

\begin{remark}
\citet{JR_vmf} argue that the choice of a threshold $t$ achieving a global $\mathrm{FAR}(t)~=~\alpha$ is not entirely relevant since it depends on the relative proportions of each sensitive attribute value $a$ within the evaluation dataset together with the relative proportion of intra-group negative pairs. They propose instead a threshold $t$ such that each group $a$ satisfies $\mathrm{FAR}_a(t) \leq \alpha$, for the max-min fairness metrics $\mathrm{FAR}_{\mathrm{min}}^\mathrm{max}(t)$ and $\mathrm{FRR}_{\mathrm{min}}^\mathrm{max}(t)$. Since we are dealing with a unique evaluation dataset, we do not use such a threshold choice, to be consistent with the other three fairness metrics (max-geomean, log-geomean and Gini) which employ a threshold $t$ such that $\mathrm{FAR}(t)=\alpha$.
\end{remark}

Other fairness metrics exist in the literature such as the maximum difference in the values $(\mathrm{FAR}_a(t))_{a \in \mathcal{A}}$ used by \citet{alasadi2019toward} and \citet{pass}. They have the disadvantage of not being normalized and are thus not interpretable, especially when comparing their values at different levels $\alpha$.


\subsection{U-statistics}\label{app:u_stat_def}

In the following, we recall the definition of a generalized $U$-statistic and show that empirical versions $\widehat{\far}_n(t)$, $\widehat{\frr}_n(t)$ of the unknown metrics $\far(t)$, $\frr(t)$ are such $U$-statistics.

\begin{definition}{\sc (Generalized $U$-statistic)}
Let $K\geq 1$ and $(d_1,\; \ldots,\; d_K)\in \mathbb{N}^{*K}$. Let $(X^{(k)}_{1},\;\ldots,\; X^{(k)}_{n_k})$, $1\leq k\leq K$, be $K$ independent samples of i.i.d. random variables, taking their values in some space $\X_k$ with distribution $F_k(dx)$ respectively. The generalized (or $K$-sample) $U$-statistic of degree $(d_1,\; \ldots,\; d_K)$ with kernel $H:\X_1^{d_1}\times \cdots \times \X_K^{d_K}\rightarrow\mathbb{R}$, square integrable with respect to the probability distribution $\mu=F_1^{\otimes d_1}\otimes \cdots \otimes F_K^{\otimes d_K}$, is defined as
\begin{equation}\label{eq:UstatG}
U_{\mathbf{n}}(H)=\frac{1}{\prod_{k=1}^K \binom{n_k}{d_k}}\sum_{I_1}\ldots\sum_{I_K} H(\mathbf{X}^{(1)}_{I_1} ; \mathbf{X}^{(2)}_{I_2}; \ldots ;\mathbf{X}^{(K)}_{I_K}),
\end{equation}
 where the symbol $\sum_{I_k}$ refers to summation over all $n_k!/(d_k!(n_k-d_k)!)$ subsets $\mathbf{X}^{(k)}_{I_k}~=~( X^{(k)}_{i_1},\;\ldots,\; X^{(k)}_{i_{d_k}})$ related to a set $I_k$ of $d_k$ indexes $1\leq i_1< \ldots <i_{d_k}\leq n_k$. It is said symmetric when $H$ is permutation symmetric in each set of $d_k$ arguments $\mathbf{X}^{(k)}_{I_k}$.
\end{definition}

Recall that $\widehat{\far}_n(t)$, $\widehat{\frr}_n(t)$ are defined, for all $t\in \mathbb{R}$, as:
\begin{eqnarray}\label{eq:emp_cdfH_bis}
\widehat{\far}_{n}(t)&=&\frac{2}{K(K-1)}\sum_{k< l}\frac{1}{n_kn_l} \sum_{i=1}^{n_k}\sum_{j=1}^{n_l}\mathbb{I}\{ s(X^{(k)}_{i},X^{(l)}_{j}) > t  \},\\
\label{eq:emp_cdfG_bis}
\widehat{\frr}_{n}(t)&=&\frac{1}{K}\sum_{k=1}^K\frac{2}{n_k(n_k-1)} \sum_{1\leq i<j\leq n_k}\mathbb{I}\{ s(X^{(k)}_{i},X^{(k)}_{j})\leq t  \}.
\end{eqnarray}

Observe that, for a fixed similarity scoring function $s$, the quantity in Equation~\ref{eq:emp_cdfG_bis} can be viewed as an average of $K$ independent (mono-sample) non-degenerate $U$-statistics of degree $2$ based on the samples $X^{(k)}_{1},\; \ldots,\; X^{(k)}_{n_k}$, $1\leq k\leq K$,  with symmetric kernel given by:
\[
g_{t}(x,\; x')=\mathbb{I}\left\{ s(x,\; x')\leq t  \right\} \quad \text{for all } (x,x')\in \X^2.
\]
Considering (\ref{eq:emp_cdfH}) now, it is a $K$-sample $U$-statistic of degree $(1,\; 1,\;\ldots,\; 1)$ with kernel given by:
\[
h_{t}(x_1,\; \ldots,\; x_K)= \frac{2}{K(K-1))}\sum_{k<l}(1-g_{t}(x_k, x_l)) \quad \text{for all } (x_1,\;\ldots,\; x_K)\in \X^K.
\]

\subsection{MORPH split}\label{app:morph_split}

In \ref{subsec:real_experiments}, we explain that the MORPH dataset is split in two parts (only for Fig.\ref{fig:motivation}).

The split was made such that each part has the same number of images ($n=27.338$) and the same identities. We discarded all identities having only one image in the original MORPH dataset, which led to remove $400$ images. Then, for identities having an even number of images, we randomly chose half of them to end up in one split, and the other half in the other split. For identities having an odd number of images, we do the same but we give an extra image to one dataset: for the next identity having an odd number of images, the extra image is given to the other dataset and we repeat the process.

\section{Additional experiments}

\subsection{Coverage of the Confidence Bands}\label{app:coverage_experiment}

\begin{wrapfigure}[17]{r}{0.46\linewidth}
\centering
\vspace{-1cm}
    \includegraphics[width=1.\linewidth]{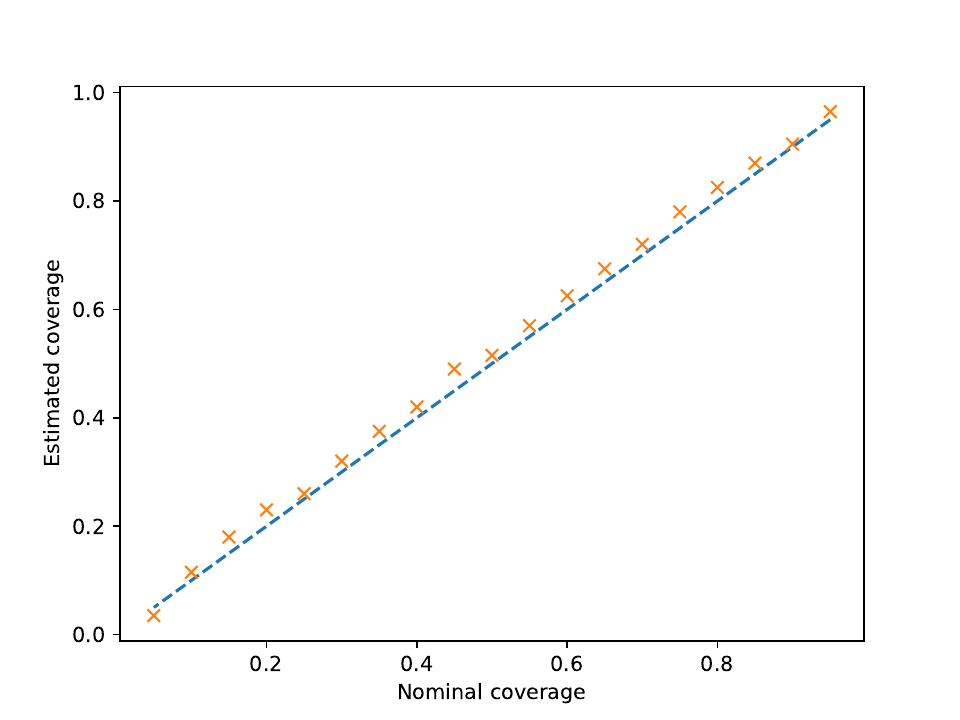}
    \caption{Estimated coverage of our confidence band method for the $\roc$ curve evaluated at $\far = 10^{-5}$ (orange). The blue dashed line represents the theoretical target.}
    \label{fig:coverage}
\end{wrapfigure}

In order to evaluate the soundness of any method to build confidence bands, it is common practice to compute the estimated coverage of the bands. Specifying the confidence level $1-\alpha_{CI}$ (also called the \textit{nominal coverage}) for our bands should lead to bands having truly a probability equal to $1-\alpha_{CI}$ to contain the true quantity $\roc(\alpha)$. To confirm whether this is the case, one may use $N_d$ synthetic datasets. On each dataset separately, the confidence bands are computed and the estimated coverage is the proportion of those $N_d$ confidence bands which contains the true $\roc$ curve. Note that this true $\roc$ curve is impossible to know with real data but may be computed/approximated with synthetic data. Ideally, the estimated coverage should be close to the nominal coverage. In this case, the method for building the confidence intervals is said to \textit{achieve nominal coverage}.

\textbf{Synthetic data.} Since our method to build confidence bands directly acts on embeddings $f(x_i) \in \mathbb{R}^p$ for a trained encoder $f$, defining the similarity $s$ (Equation~\ref{eq:cos_sim}), the simplest way to create synthetic datasets is to generate synthetic embeddings. It turns out that a natural statistical model for FR embeddings is the mixture of von Mises-Fisher (vMF) distributions, where each component of the mixture is associated to one identity of the dataset (see \textit{e.g.} \cite{gentric_vMF} or \cite{JR_vmf}). In details, the embeddings sharing a same identity are modelled as realizations of one vMF distribution, a gaussian in dimension $p$ projected onto the unit hypersphere of dimension $p$, characterized by its centroid (its mean vector) and its concentration parameter (the inverse of the variance). We have randomly drawn $K=10^3$ centroids on a hypersphere of dimension $p=128$ using the Marsaglia method \citep{Marsaglia1972ChoosingAP}. The concentration parameter of each vMF distribution is drawn uniformly in $[100, 800]$. Those centroids and concentration parameters fully define our $K$ synthetic identities. A synthetic dataset is generated by sampling $n_k=10$ points from the vMF distribution \citep{kim2021pytorch} associated to identity $k$, for each identity $k$ within the $K$ identities. We obtain a synthetic dataset of $n = \sum_{k=1}^K n_k =10^4$ embeddings. The process is repeated to generate $N_d = 200$ synthetic datasets, sharing the same $K$ identities (\textit{i.e.} the centroids and concentration parameters are shared by all datasets). 

\textbf{Protocol.} For each dataset, we use our recentered bootstrap method (with $B=200$ bootstrap samples) to compute confidence intervals, at a given nominal coverage $1-\alpha_{CI}$, for the ROC curve evaluated at a $\far$ level equal to $\alpha = 10^{-5}$. This process results in $N_d$ confidence intervals. To approximate the ground-truth ROC value which those confidence intervals should contain, we concatenate all $N_d$ datasets into one dataset of $N_d \times n = 2 \times 10^6$ embeddings and find its empirical ROC curve $\widehat{\roc}_n(\alpha)$, still evaluated at a $\far$ level equal to $\alpha = 10^{-5}$. The estimated coverage is the proportion of the $N_d$ confidence intervals which contains this ground-truth $\roc$. The result is displayed in Figure~\ref{fig:coverage}, for several values of nominal coverage. We list the values displayed on Fig.\ref{fig:coverage} in Table~\ref{tab:coverage}. The recentered bootstrap nearly achieves perfect nominal coverage, which supports the soundness of the confidence bands presented within the paper. In other words, the width of each confidence interval is nearly equal to the theoretical target. Theorem~\ref{thm:asym_coverage} claims that this fact is true when the number of evaluation data goes to infinity. In the present case, we demonstrate that the result holds even in the non asymptotic regime. 

\begin{table}[h!]
\begin{center}
\captionsetup{width=.7\linewidth}
\caption{Estimated coverage of the $\roc$ curve evaluated at $\far~=~10^{-5}$, using the recentered bootstrap, for several nominal coverage values.}
\label{tab:coverage}
\begin{tabular}{ |c|c| } 
 \hline
 Nominal coverage & Estimated coverage \\ 
 \hline
 0.95 & 0.96  \\ 
 0.90 & 0.90  \\
 0.85 & 0.87  \\ 
 0.80 & 0.82  \\
 0.75 & 0.78  \\ 
 0.70 & 0.72  \\
 0.65 & 0.67  \\ 
 0.60 & 0.62  \\
 0.55 & 0.57  \\ 
 0.50 & 0.51  \\
 0.45 & 0.49  \\ 
 0.40 & 0.42  \\
 0.35 & 0.37  \\ 
 0.30 & 0.32  \\
 0.25 & 0.26  \\ 
 0.20 & 0.23  \\
 0.15 & 0.18  \\ 
 0.10 & 0.11  \\
 0.05 & 0.04  \\ 
 \hline
\end{tabular}
\end{center}
\end{table}

\begin{figure}[h!]
\centering
\begin{minipage}{.46\textwidth}
  \centering
\includegraphics[width=1.05\linewidth]{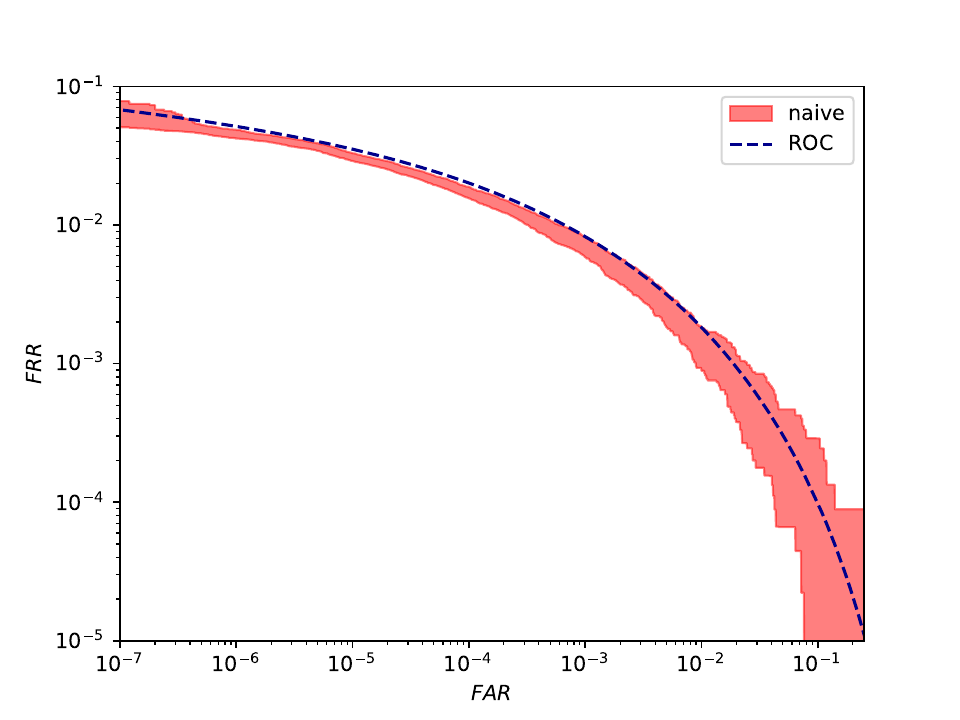}
  \captionof{figure}{Confidence bands obtained by the naive bootstrap for the $\roc$ curve on one of the $N_d$ synthetic datasets. The ground-truth $\roc$ curve is depicted in dashed lines.}
  \label{fig:synthetic_naive_ci}
\end{minipage}%
\hspace{0.5cm}
\begin{minipage}{.46\textwidth}
  \centering
\includegraphics[width=1.05\linewidth]{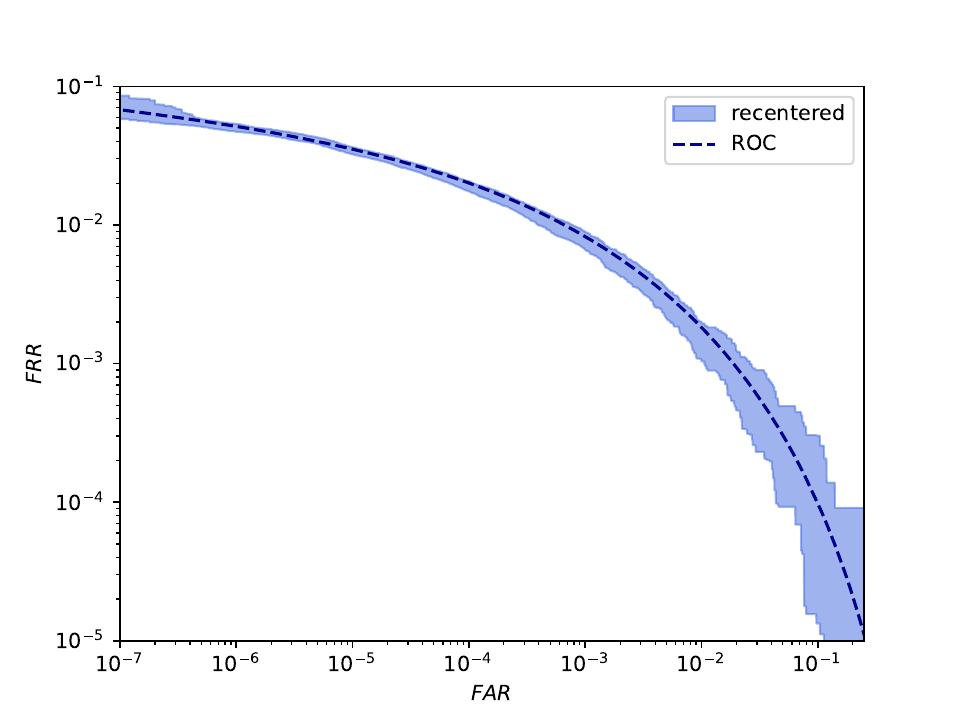}
  \captionof{figure}{Confidence bands obtained by the recentered bootstrap for the $\roc$ curve on one of the $N_d$ synthetic datasets. The ground-truth $\roc$ curve is depicted in dashed lines.}
  \label{fig:synthetic_recentered_ci}
\end{minipage}
\end{figure}

\begin{figure}
      \centering
      \includegraphics[width=0.5\linewidth]{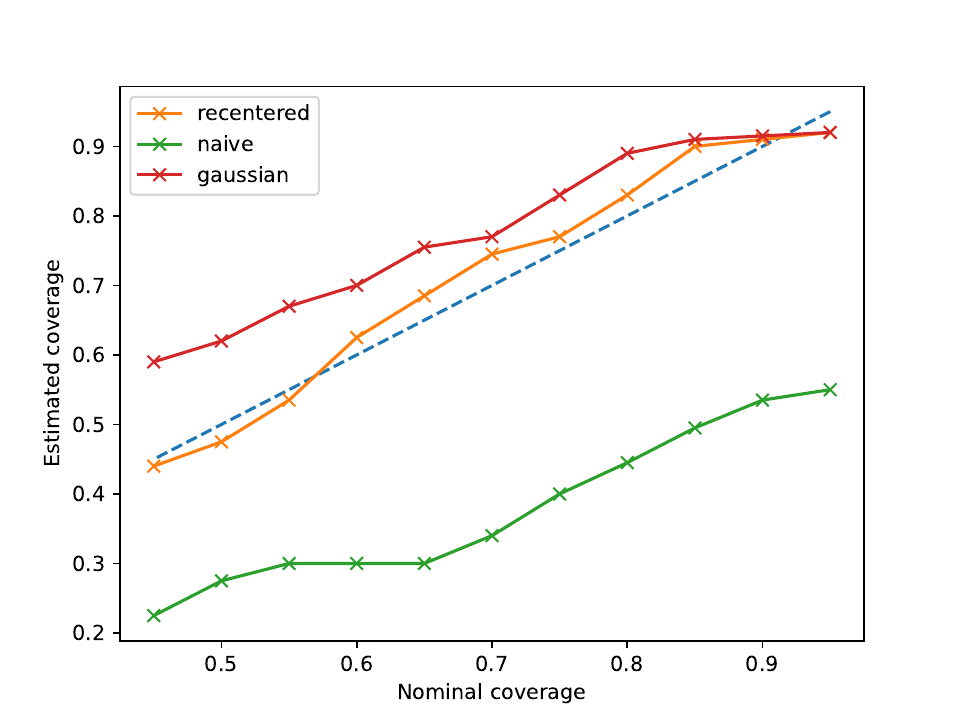}
      \caption{Estimated coverage of three confidence band methods for the $\roc$ curve evaluated at $\far = 10^{-1}$. The blue dashed line represents the theoretical target. Three methods to build a confidence interval are depicted: the recentered bootstrap (orange), the naive bootstrap (green) and a gaussian approximation of the recentered bootstrap (red).}
      \label{fig:coverage_baseline}
  \end{figure}

\textbf{Naive bootstrap.} We now compute the estimated coverage of the $\roc$ curve, when using the naive bootstrap \citep{Bertail1}, as a baseline for our method. Computing confidence bands, at nominal coverage $1-\alpha_{CI}=95\%$, for the $\roc$ curve on one of the $N_d$ datasets leads to Fig.~\ref{fig:synthetic_naive_ci} (using the naive bootstrap) and Fig.~\ref{fig:synthetic_recentered_ci} (using the recentered bootstrap). As seen with real data in Fig.\ref{fig:naive_recentered_ci}, the naive bootstrap underestimates the $\roc$ curve, and thus the confidence bands. In Fig.\ref{fig:synthetic_naive_ci}-\ref{fig:synthetic_recentered_ci}, the ground-truth $\roc$ curve (not the empirical one) is depicted in dashed lines. 
Note that the confidence band from the naive bootstrap does not contain the ground-truth $\roc$ for many $\far$ values. This leads to an estimated coverage for the naive bootstrap that is experimentally equal to zero, no matter what is the confidence level/nominal coverage. For a real comparison between the naive and recentered bootstraps, we choose a $\far$ value for which the naive bootstrap has chances of being accurate. In this sense, we do not consider the $\roc$ evaluated at $\far=10^{-5}$, as in Fig.\ref{fig:coverage}, but rather at $\far=10^{-1}$, where the ground-truth $\roc$ seems to be contained within the bands of the naive bootstrap (see Fig.~\ref{fig:synthetic_naive_ci}). We employ the same protocol to compute the estimated coverage as for Fig.~\ref{fig:coverage}. The results are presented in Fig.~\ref{fig:coverage_baseline} for the naive and the recentered bootstraps in particular, with nominal coverage values leading to a non-zero estimated coverage for the naive bootstrap. We list the values displayed on Fig.\ref{fig:coverage_baseline} in Table~\ref{tab:coverage_baseline}. Note that the naive bootstrap \citep{Bertail1} is far from achieving nominal coverage. This underlines the necessity of the recentering step in the bootstrap methodology. 

\begin{table}[h!]
\begin{center}
\caption{Estimated coverage of the $\roc$ curve evaluated at $\far~=~10^{-1}$, using three confidence band methods, for several nominal coverage values. For each nominal coverage, the best method is displayed in bold, the second best one is underlined.}
\label{tab:coverage_baseline}
\begin{tabular}{ c||c|c|c } 
&\multicolumn{3}{|c|}{Estimated coverage} \\
 \hline
 \multicolumn{1}{|c||}{Nominal coverage} & Recentered bootstrap &Naive bootstrap& \multicolumn{1}{|c|}{Gaussian approximation} \\ 
 \hline
\multicolumn{1}{|c||}{0.95} & \textbf{0.92} & 0.55 & \multicolumn{1}{|c|}{\textbf{0.92}} \\
\multicolumn{1}{|c||}{0.90} & \textbf{0.91} & 0.54 & \multicolumn{1}{|c|}{\underline{0.92}} \\
\multicolumn{1}{|c||}{0.85} & \textbf{0.90} & 0.50 & \multicolumn{1}{|c|}{\underline{0.91}} \\
\multicolumn{1}{|c||}{0.80} & \textbf{0.83} & 0.45 & \multicolumn{1}{|c|}{\underline{0.89}} \\
\multicolumn{1}{|c||}{0.75} & \textbf{0.77} & 0.40 & \multicolumn{1}{|c|}{\underline{0.83}} \\
\multicolumn{1}{|c||}{0.70} & \textbf{0.74} & 0.34 & \multicolumn{1}{|c|}{\underline{0.77}} \\
\multicolumn{1}{|c||}{0.65} & \textbf{0.68} & 0.30 & \multicolumn{1}{|c|}{\underline{0.76}} \\
\multicolumn{1}{|c||}{0.60} & \textbf{0.62} & 0.30 & \multicolumn{1}{|c|}{\underline{0.70}} \\
\multicolumn{1}{|c||}{0.55} & \textbf{0.53} & 0.30 & \multicolumn{1}{|c|}{\underline{0.67}} \\
\multicolumn{1}{|c||}{0.50} & \textbf{0.48} & 0.28 & \multicolumn{1}{|c|}{\underline{0.62}} \\
\multicolumn{1}{|c||}{0.45} & \textbf{0.44} & 0.23 & \multicolumn{1}{|c|}{\underline{0.59}} \\
 \hline
\end{tabular}
\end{center}
\end{table}

\textbf{Gaussian approximation.} In order to provide a more meaningful baseline than the naive bootstrap \citep{Bertail1}, we design in the following another confidence band method. This method is significantly close to the recentered bootstrap, as there is no such method in the literature for similarity scoring tasks. The process is almost the same than for Algorithm~\ref{alg:CI} (recentered bootstrap) where one forms bootstrap samples, then computes a bootstrap version $\widehat{\roc}_n^*$. Using the recentering $\widetilde{\roc}_n$, one gets recentered bootstrap versions of the $\roc$ curve, defined as $\widehat{\roc}_n(\alpha) + \hat{\epsilon}_n^{(2)}(\alpha)$ 
where $\hat{\epsilon}_n^{(2)}(\alpha) =\widehat{\roc}_n^*(\alpha)-\widetilde{\roc}_n(\alpha)$ (see \ref{subsec:bootstrap}). Algorithm~\ref{alg:CI} then defines the confidence bands as the $\frac{\alpha_{CI}}{2}$-th and $\frac{1-\alpha_{CI}}{2}$-th quantiles of those recentered bootstrap versions of the $\roc$. Instead of considering those quantiles, a simple idea would be to model $\widehat{\roc}_n(\alpha) + \hat{\epsilon}_n^{(2)}(\alpha)$ as a gaussian random variable. As for Algorithm~\ref{alg:CI} where one simulates $B$ realizations of $\hat{\epsilon}_n^{(2)}(\alpha)$, the same process is applied here to estimate the parameters (mean and variance) of the gaussian distribution of $\widehat{\roc}_n(\alpha) + \hat{\epsilon}_n^{(2)}(\alpha)$. Once the parameters estimated, this gaussian approximation allows us to define the bounds of the confidence interval as the $\frac{\alpha_{CI}}{2}$-th and $\frac{1-\alpha_{CI}}{2}$-th quantiles of the corresponding gaussian law. Note that this approximation relies a lot on the recentering which we present within the paper. The estimated coverage of the $\roc$ curve using this gaussian approximation is displayed in Fig.~\ref{fig:coverage_baseline} and Table~\ref{tab:coverage_baseline}. This baseline outperforms significantly the naive bootstrap \citep{Bertail1}, while being slightly less accurate than our recentered bootstrap. A disadvantage of this method, compared to the recentered bootstrap, is that the confidence intervals are too wide, as seen in Fig.~\ref{fig:coverage_baseline} (the estimated coverage in red is most of the time higher than the nominal coverage). Having too wide confidence intervals has its flaws, as one could claim that two models are indistinguishable in terms of performance ($\roc$) whereas they are not. Confidence bands must be precise, proven valid, in order to be meaningful. 

\subsection{Comparison of the uncertainty of fairness metrics}\label{app:normalized_uncertainty_fairness_expes}

\begin{figure}
\centering
\begin{minipage}{.46\textwidth}
  \centering
  \includegraphics[width=1.05\linewidth]{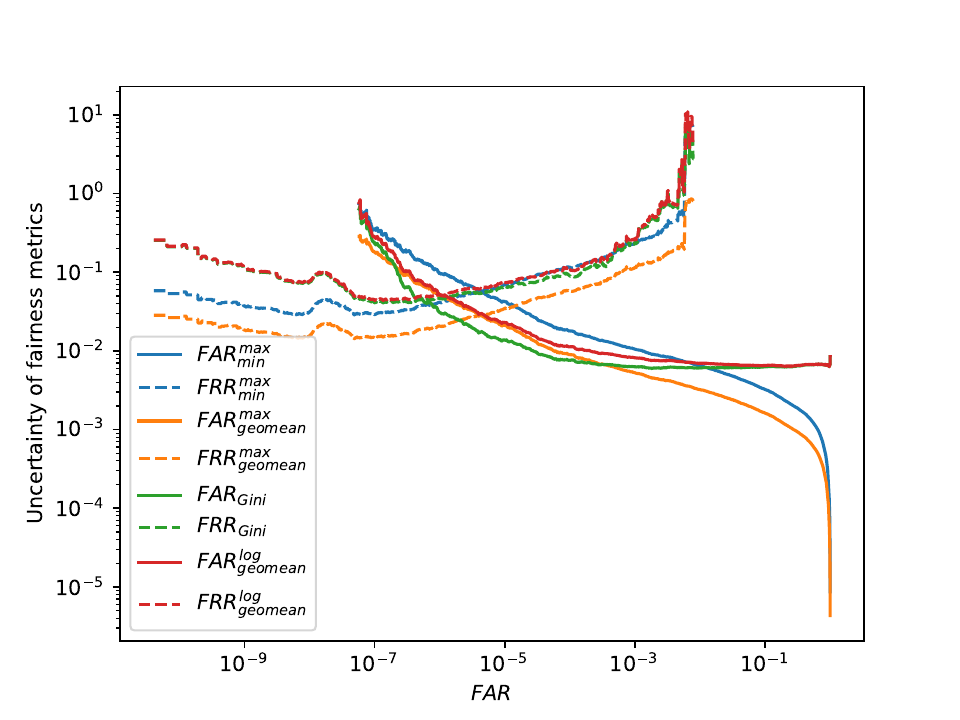}
  \captionof{figure}{Normalized uncertainty of several fairness metrics ($\far$ fairness in solid lines, $\frr$ fairness in dashed lines). The dataset is MORPH, the sensitive attribute is the gender label and the encoder $f$ is CurricularFace.}
  \label{fig:std_fairness_curri_g}
\end{minipage}%
\hspace{0.5cm}
\begin{minipage}{.46\textwidth}
  \centering
  \includegraphics[width=1.05\linewidth]{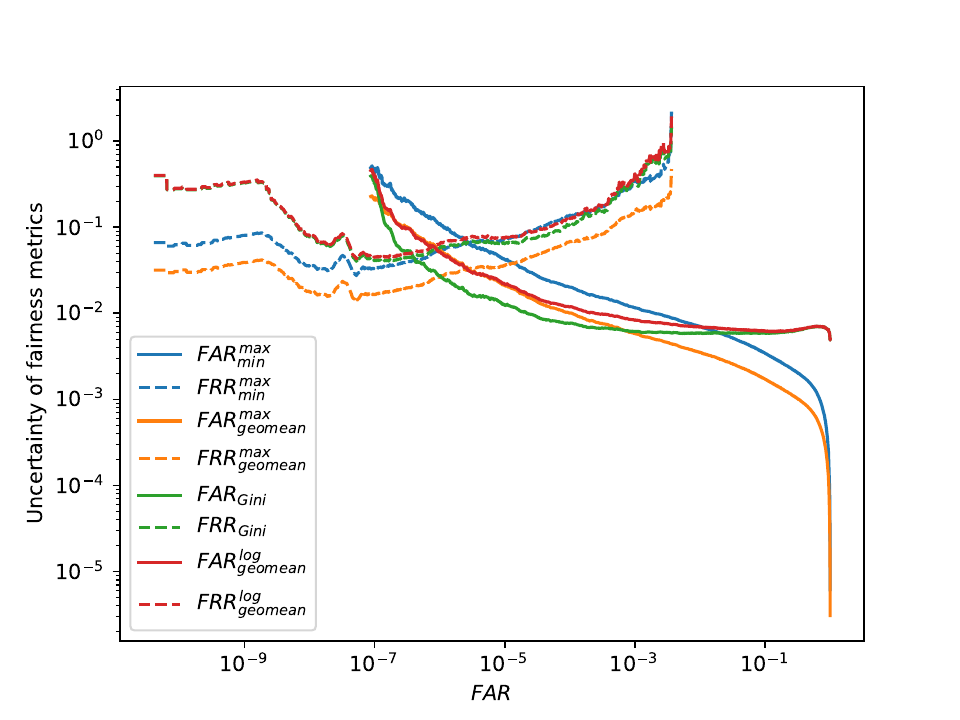}
  \captionof{figure}{Normalized uncertainty of several fairness metrics ($\far$ fairness in solid lines, $\frr$ fairness in dashed lines). The dataset is MORPH, the sensitive attribute is the gender label and the encoder $f$ is CosFace.}
  \label{fig:std_fairness_cos_g}
\end{minipage}
\end{figure}

In this section, we compute the normalized uncertainty (Eq.~\ref{eq:normalized_uncertainty}) for the considered FR fairness metrics, as for Fig.\ref{fig:std_fairness}. In Fig.~\ref{fig:std_fairness_curri_g} and \ref{fig:std_fairness_cos_g}, we display those uncertainty metrics computed on the MORPH dataset, with the gender label as the sensitive attribute, for two distinct models (CurricularFace and CosFace).

\begin{figure}
\centering
\begin{minipage}{.46\textwidth}
  \centering
  \includegraphics[width=1.05\linewidth]{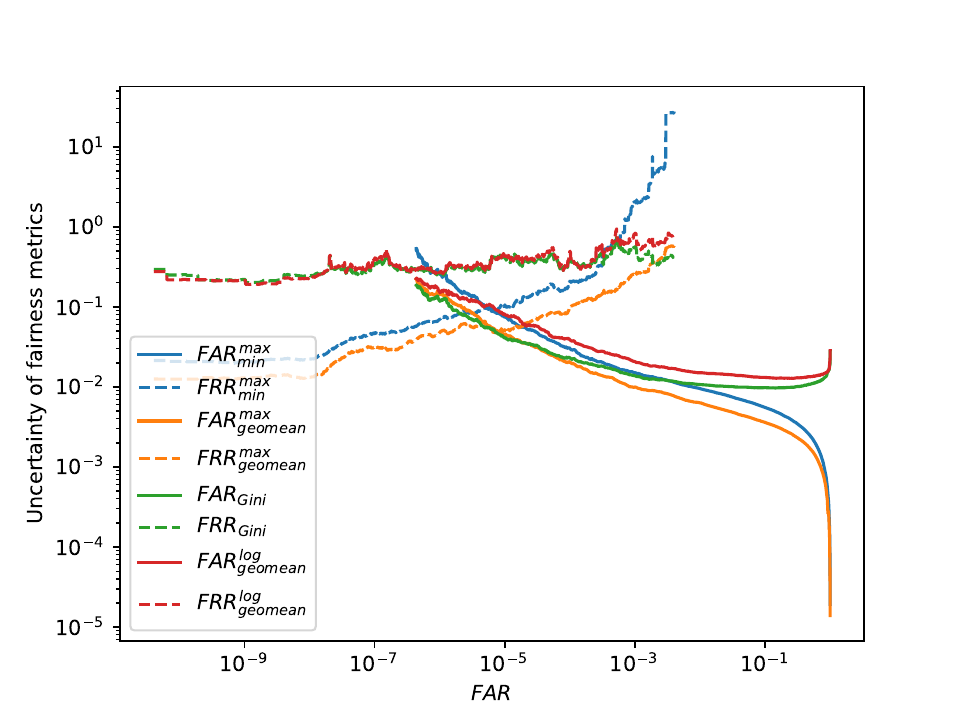}
  \captionof{figure}{Normalized uncertainty of several fairness metrics ($\far$ fairness in solid lines, $\frr$ fairness in dashed lines). The dataset is MORPH, the sensitive attribute is the age label and the encoder $f$ is CurricularFace.}
  \label{fig:std_fairness_curri_a}
\end{minipage}%
\hspace{0.5cm}
\begin{minipage}{.46\textwidth}
  \centering
  \includegraphics[width=1.05\linewidth]{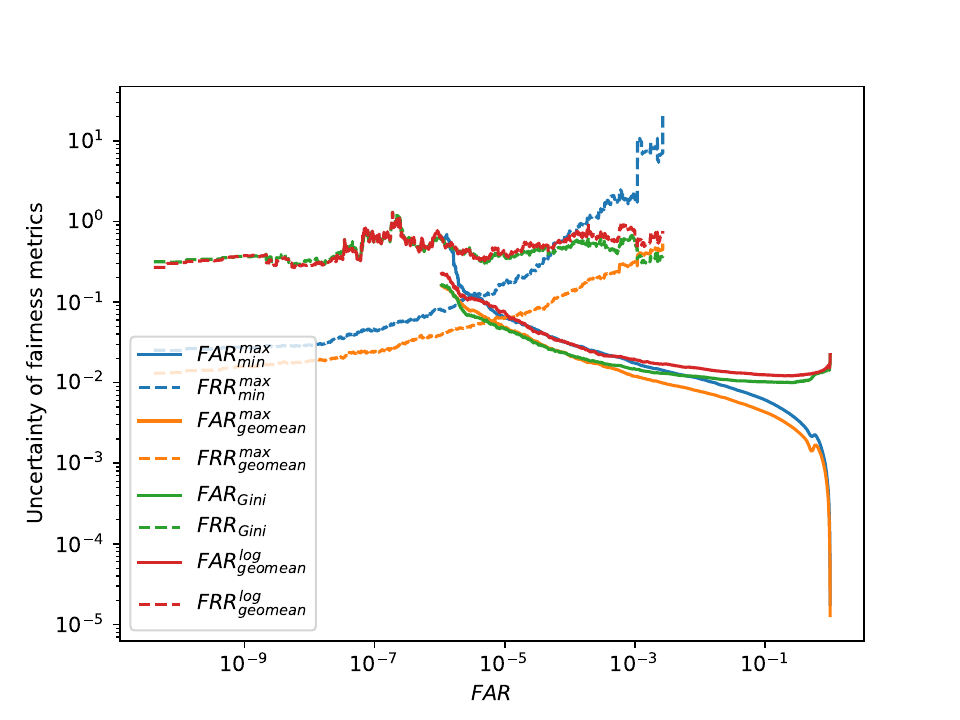}
  \captionof{figure}{Normalized uncertainty of several fairness metrics ($\far$ fairness in solid lines, $\frr$ fairness in dashed lines). The dataset is MORPH, the sensitive attribute is the age label and the encoder $f$ is CosFace.}
  \label{fig:std_fairness_cos_a}
\end{minipage}
\end{figure}

In Fig.~\ref{fig:std_fairness_curri_a} and \ref{fig:std_fairness_cos_a}, we display those uncertainty metrics computed on the MORPH dataset, with the age label as the sensitive attribute, for two distinct models (CurricularFace and CosFace). The discrete age labels provided by the MORPH dataset are age group labels: '<20', '20-29', '30-39', '40-49', '50+'.

\begin{figure}
      \centering
      \includegraphics[width=0.5\linewidth]{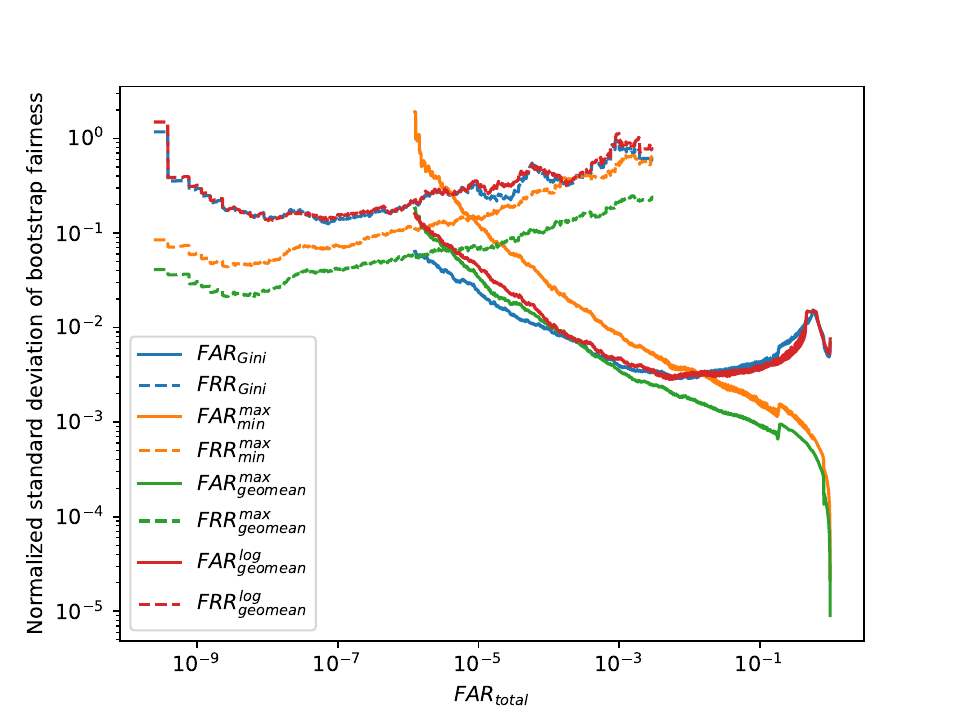}
      \caption{Normalized uncertainty of several fairness metrics ($\far$ fairness in solid lines, $\frr$ fairness in dashed lines). The dataset is RFW, the sensitive attribute is the race label and the encoder $f$ is ArcFace with ResNet backbone.}
      \label{fig:std_fairness_rfw_arc}
  \end{figure}

Lastly, we change the evaluation dataset and the backbone of the pre-trained model. We use as encoder the trained\footnote{\url{https://github.com/deepinsight/insightface/tree/master/recognition/arcface_torch}.} model ArcFace \citep{arcface} whose CNN architecture is a ResNet100 \citep{resnet100_forFR}. It has been trained on the MS1M-RetinaFace dataset, introduced by \citep{ms1m_retinaface} in the ICCV 2019 Lightweight Face Recognition Challenge. We choose the dataset RFW \citep{RFW} as evaluation dataset. It is composed of $40$k face images from $11$k distinct identities. This dataset is also provided with ground-truth race labels (the four available labels are: African, Asian, Caucasian, Indian) and is widely used for fairness evaluation since it is equally distributed among the race subgroups, in terms of images and identities. The official RFW protocol only considers a few matching pairs among all the possible pairs given the whole RFW dataset. The number of images is typically not enough to get good estimates of our fairness metrics at low $\mathrm{FAR}$. To overcome this, we consider all possible same-race matching pairs among the whole RFW dataset. In Fig.\ref{fig:std_fairness_rfw_arc}, we use ArcFace with ResNet100 backbone evaluated on RFW with the sensitive attribute being the race label.

\subsection{Other fairness metrics for Figure~\ref{fig:method_decision_fairness} }\label{app:fig4_analogues}

In Fig.~\ref{fig:method_decision_fairness}, we provide a use-case of model selection depending on a strict fairness constraint. The uncertainty of the fairness metric $\frr_{\mathrm{min}}^{\mathrm{max}}$ can be high for some models as it can be low. A strict fairness constraint may forbid some models, given their high fairness uncertainty.

\begin{figure}
\centering
\begin{minipage}{.46\textwidth}
  \centering
  \includegraphics[width=1.05\linewidth]{imgs_iclr/methodology_frr_fairness_decision.pdf}
  \captionof{figure}{Confidence bands at $95$\% confidence level for the $\mathrm{FRR}_{\mathrm{min}}^\mathrm{max}$ fairness metric, for two models (ArcFace, AdaCos). The empirical fairness metrics are depicted as solid lines.}
\label{fig:max_min_methodology_frr_fairness_decision}
\end{minipage}%
\hspace{0.5cm}
\begin{minipage}{.46\textwidth}
  \centering
  \includegraphics[width=1.05\linewidth]{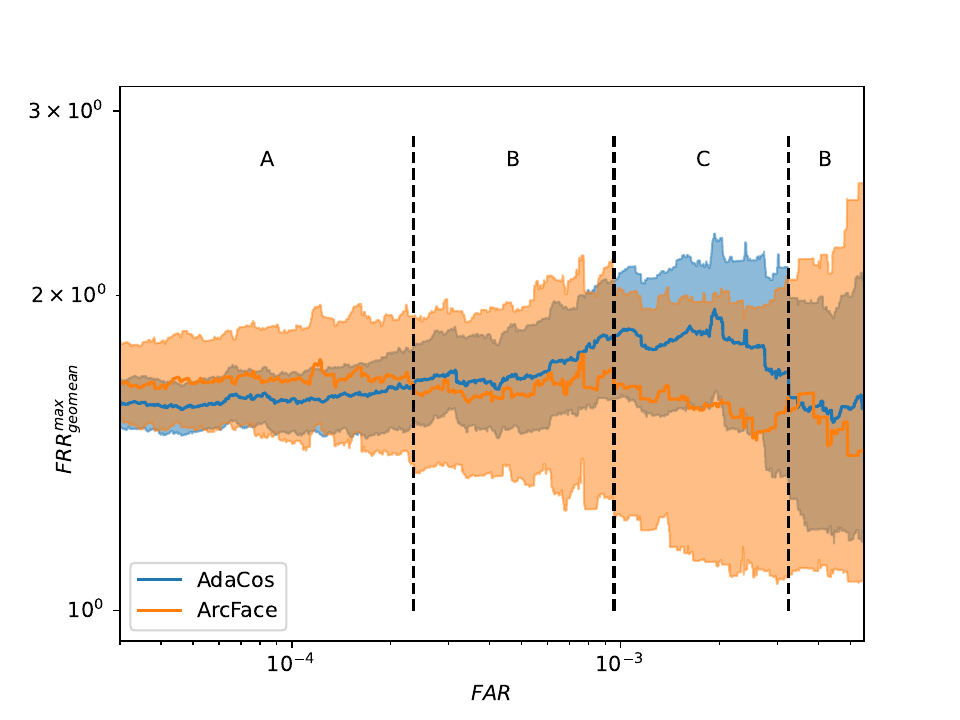}
  \captionof{figure}{Confidence bands at $95$\% confidence level for the $\mathrm{FRR}_{\mathrm{geomean}}^\mathrm{max}$ fairness metric, for two models (ArcFace, AdaCos). The empirical fairness metrics are depicted as solid lines.}
\label{fig:max_geomean_methodology_frr_fairness_decision}
\end{minipage}
\end{figure}

\begin{figure}
\centering
\begin{minipage}{.46\textwidth}
  \centering
  \includegraphics[width=1.05\linewidth]{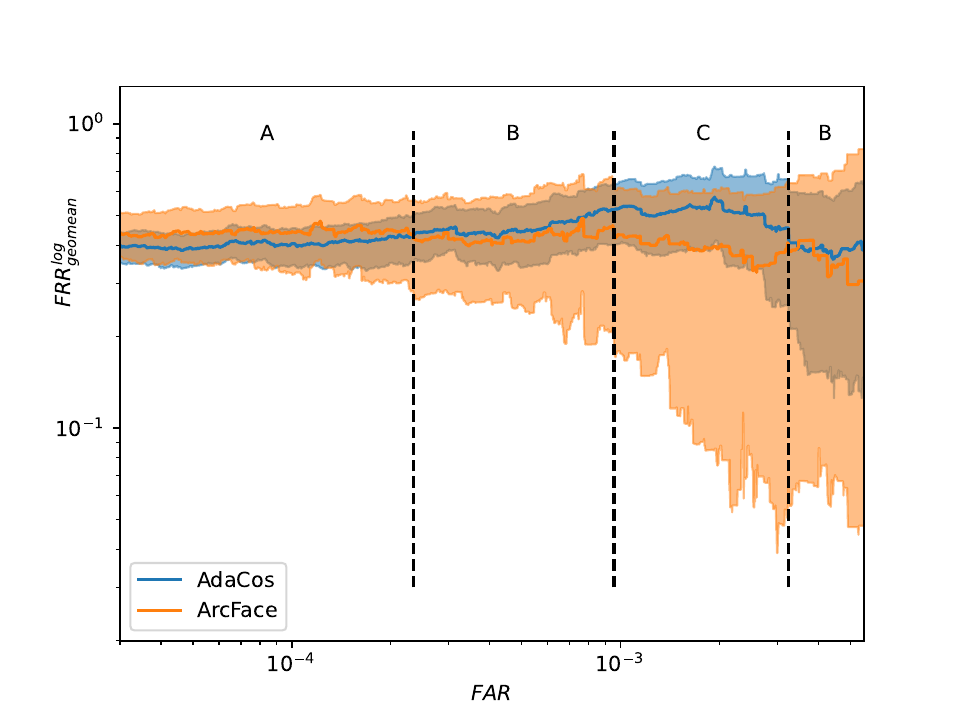}
  \captionof{figure}{Confidence bands at $95$\% confidence level for the $\mathrm{FRR}_{\mathrm{geomean}}^\mathrm{log}$ fairness metric, for two models (ArcFace, AdaCos). The empirical fairness metrics are depicted as solid lines.}
\label{fig:log_geomean_methodology_frr_fairness_decision}
\end{minipage}%
\hspace{0.5cm}
\begin{minipage}{.46\textwidth}
  \centering
  \includegraphics[width=1.05\linewidth]{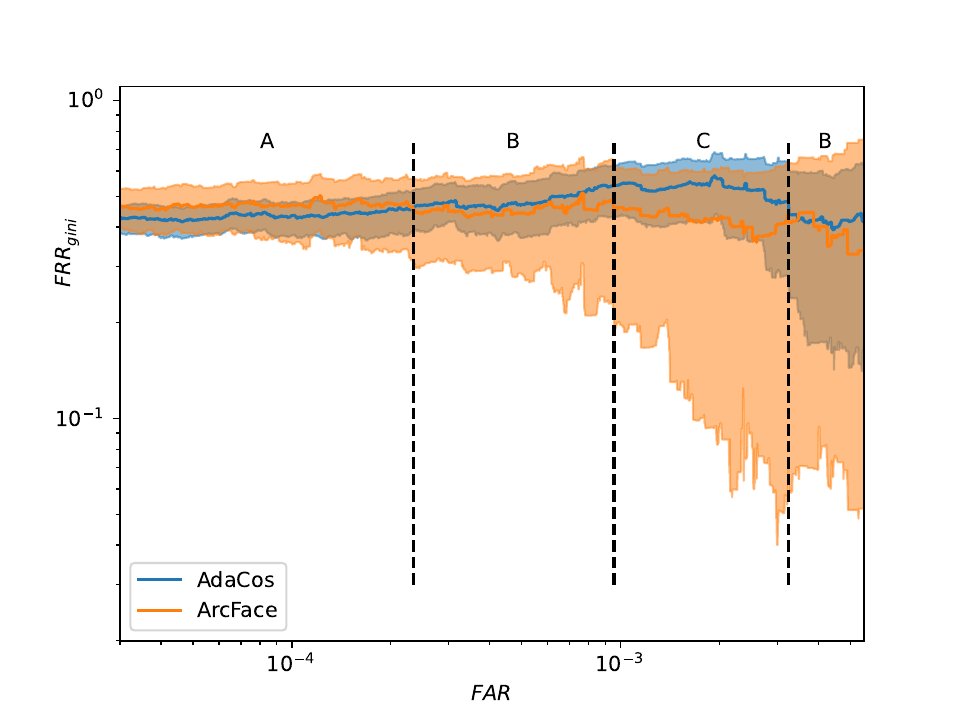}
  \captionof{figure}{Confidence bands at $95$\% confidence level for the $\mathrm{FRR}_{\mathrm{Gini}}^\mathrm{}$ fairness metric, for two models (ArcFace, AdaCos). The empirical fairness metrics are depicted as solid lines.}
\label{fig:gini_methodology_frr_fairness_decision}
\end{minipage}
\end{figure}

We illustrated this use-case with the fairness metric $\frr_{\mathrm{min}}^{\mathrm{max}}$. In this section, we provide the equivalent of Fig.~\ref{fig:method_decision_fairness}, but for all $\frr$ fairness metrics, for completeness. Those equivalents are displayed in Fig.~\ref{fig:max_geomean_methodology_frr_fairness_decision} ($\mathrm{FRR}_{\mathrm{geomean}}^\mathrm{max}$), Fig.~\ref{fig:log_geomean_methodology_frr_fairness_decision} ($\mathrm{FRR}_{\mathrm{geomean}}^\mathrm{log}$) and Fig.~\ref{fig:gini_methodology_frr_fairness_decision} ($\mathrm{FRR}_{\mathrm{Gini}}^\mathrm{}$). As for Fig.~\ref{fig:method_decision_fairness}, the evaluation dataset is MORPH, the sensitive attribute is the gender label and the number of bootstrap samples given to Algorithm~\ref{alg:CI} is $B=200$.

Note that the conclusions from Fig.~\ref{fig:method_decision_fairness} are unchanged when considering other fairness metrics. Indeed, the fairness metrics of AdaCos and ArcFace all exhibit exactly the same behaviour : both empirical fairness metrics intersect at the same $\far$ level for all fairness metrics, and the upper bounds of both fairness metrics also intersect at the same $\far$ levels for all fairness metrics. This striking fact highlights the fact that all fairness metrics measure the same performance differentials. However, one significant difference between those metrics is their uncertainty: the width of the confidence bands can be high, relatively to the value of the empirical fairness, making the fairness metric not so robust. The comparison of the uncertainty between fairness metrics is displayed in Fig.~\ref{fig:std_fairness} and in Section~\ref{app:normalized_uncertainty_fairness_expes}.

\subsection{Fairness metrics on RFW dataset}

\begin{figure}
\centering

\begin{minipage}{.46\textwidth}
  \centering
\includegraphics[width=1.05\linewidth]{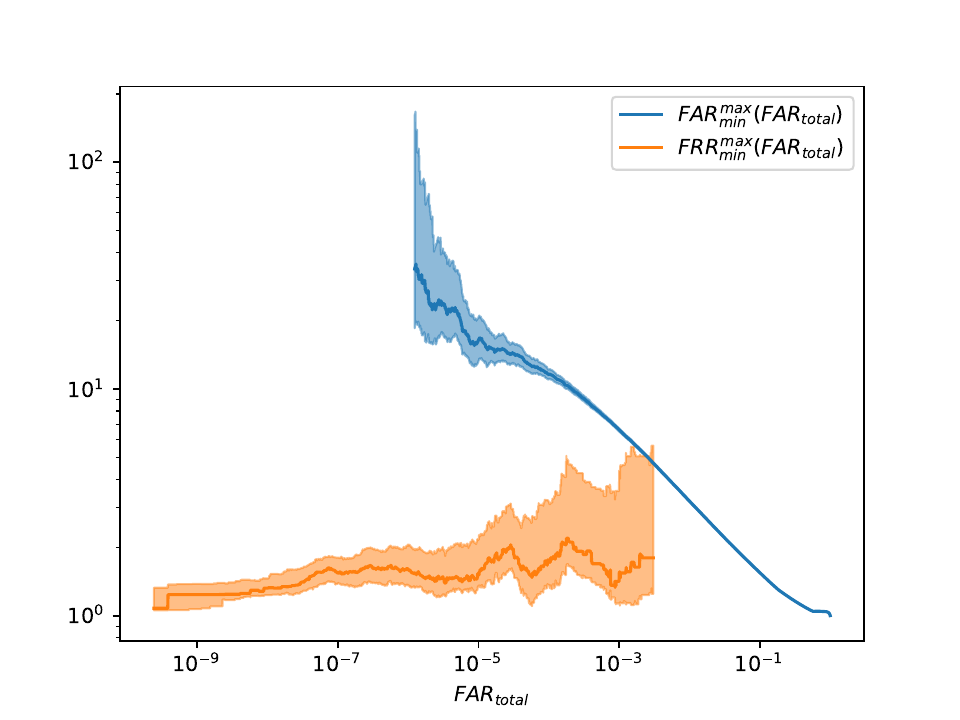}
  \captionof{figure}{Confidence bands at $95$\% confidence level for the $\mathrm{FAR}_{\mathrm{min}}^\mathrm{max}$ and $\mathrm{FRR}_{\mathrm{min}}^\mathrm{max}$ fairness metrics. $B=100$ bootstrap samples are used. The empirical fairness metrics are depicted as solid lines.}
  \label{fig:rfw_max_min_ci}
\end{minipage}
\hspace{0.5cm}
\begin{minipage}{.46\textwidth}
  \centering
\includegraphics[width=1.05\linewidth]{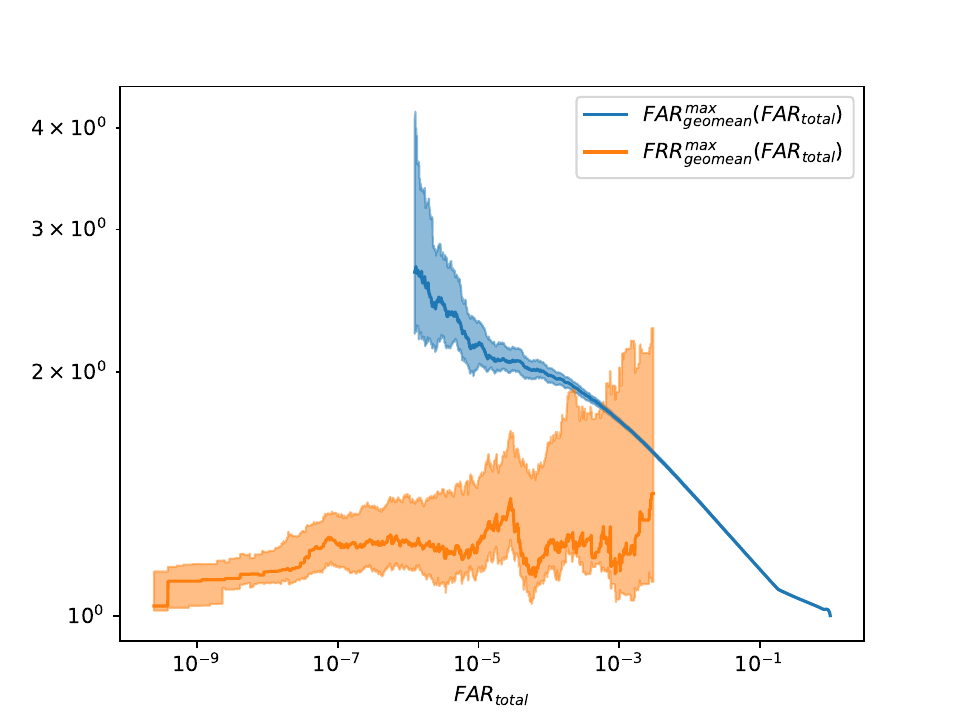}
  \captionof{figure}{Confidence bands at $95$\% confidence level for the $\mathrm{FAR}_{\mathrm{geomean}}^\mathrm{max}$ and $\mathrm{FRR}_{\mathrm{geomean}}^\mathrm{max}$ fairness metrics. $B=100$ bootstrap samples are used. The empirical fairness metrics are depicted as solid lines.}
  \label{fig:rfw_max_geomean_ci_bis}
\end{minipage}%

\begin{minipage}{.46\textwidth}
  \centering
\includegraphics[width=1.05\linewidth]{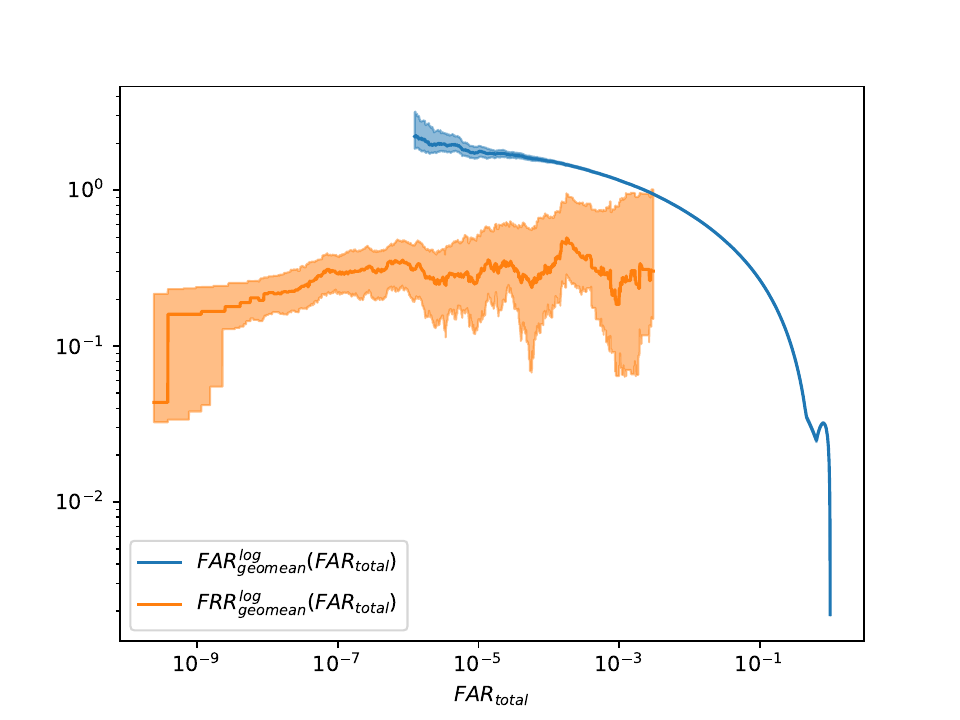}
  \captionof{figure}{Confidence bands at $95$\% confidence level for the $\mathrm{FAR}_{\mathrm{geomean}}^\mathrm{log}$ and $\mathrm{FRR}_{\mathrm{geomean}}^\mathrm{log}$ fairness metrics. $B=100$ bootstrap samples are used. The empirical fairness metrics are depicted as solid lines.}
  \label{fig:rfw_log_geomean_ci}
\end{minipage}
\hspace{0.5cm}
\begin{minipage}{.46\textwidth}
  \centering
\includegraphics[width=1.05\linewidth]{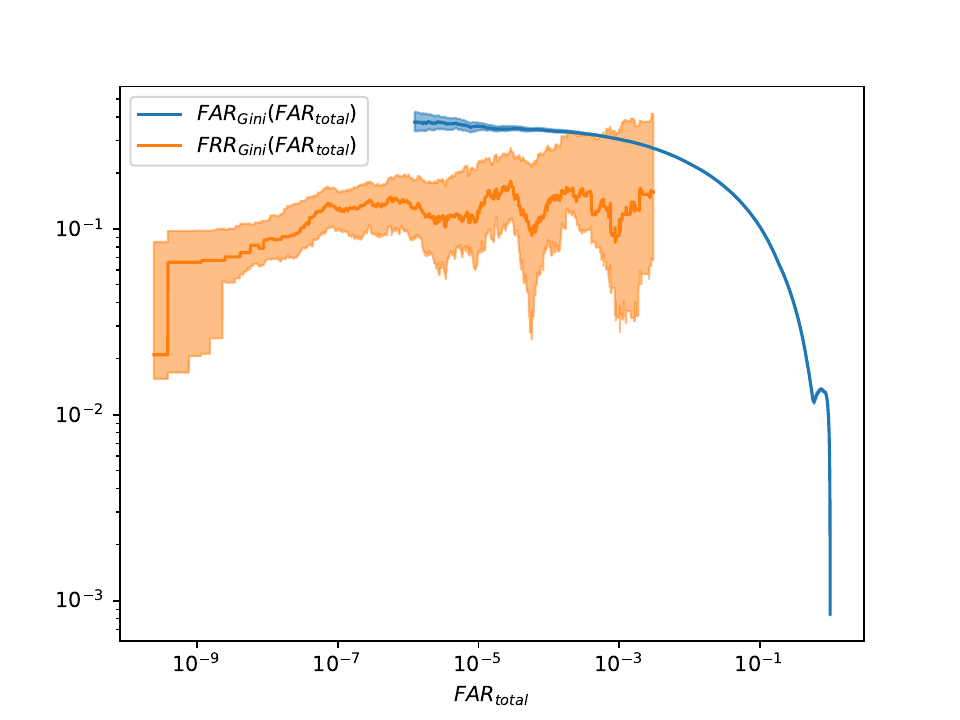}
  \captionof{figure}{Confidence bands at $95$\% confidence level for the $\mathrm{FAR}_{\mathrm{Gini}}$ and $\mathrm{FRR}_{\mathrm{Gini}}$ fairness metrics. $B=100$ bootstrap samples are used. The empirical fairness metrics are depicted as solid lines.}
  \label{fig:rfw_gini_ci}
\end{minipage}%
\end{figure}

For comprehensiveness, we give the confidence bands for all eight considered fairness metrics, listed in section~\ref{app:fairness_metrics} (see Figures~\ref{fig:rfw_max_min_ci}, \ref{fig:rfw_max_geomean_ci_bis}, \ref{fig:rfw_log_geomean_ci}, \ref{fig:rfw_gini_ci}). This is the result of Algorithm~\ref{alg:CI} applied to each fairness metric. We chose the RFW dataset~\citep{RFW}, as it is balanced in race labels. We already used the gender and age labels within the paper. The model is ArcFace and we evaluate its fairness at each $\far_{\mathrm{total}} = \alpha \in [0,1]$ level.

\section{Pseudo-code}\label{app:code}

\subsection{Pseudo-code for the naive bootstrap}\label{subsec:code_naive_bootstrap}

Algorithm~\ref{alg:naive_ROC} provides a pseudo-code for the naive bootstrap of $\widehat{\mathrm{ROC}}_n(\alpha)$, described in section~\ref{subsec:bootstrap}. The idea is to sample with replacement $n_k$ images among the images of identity $k$ (the i.i.d. variables $X^{(k)*}_{1},\; \ldots,\; X^{(k)*}_{n_k}$ with distribution $\hat{F}_k$) for each identity $k = 1, \dotsc, K$. With those new data, one is able to compute $\widehat{\far}^*_{n}(t)$ and $\widehat{\frr}^*_{n}(t)$ from Eq.~\ref{eq:boot_far}~and~\ref{eq:boot_frr}, and thus a bootstrap version $\widehat{\roc}_n^*(\alpha)$ of the ROC curve.

By repeating this process $B$ times, one is able to better estimate the variability of $\widehat{\mathrm{ROC}}_n(\alpha)$. At each step $b$, for $b = 1, \dotsc, B$, one creates a bootstrap sample $X_{(b)}$ of image data and ends up with a bootstrap version $\widehat{\mathrm{ROC}}_{n, (b)}^*(\alpha)$ of the empirical $\widehat{\mathrm{ROC}}_n(\alpha)$. Accumulating those values for many steps $b$ and for all $\alpha \in (0,1)$ gives the bundle of light-blue ROC curves in Fig.~\ref{fig:roc_bootstrap}. As explained in section~\ref{subsec:bootstrap}, those curves are not centered around the empirical $\widehat{\mathrm{ROC}}_n$, but around its V-statistic counterpart given by $\widetilde{\mathrm{ROC}}_n(\alpha) =\widetilde{\frr}_{n}\circ (\widehat{\far}_{n})^{-1}(\alpha)$.

\algnewcommand\algorithmicinput{\textbf{Input:}}
\algnewcommand\Input{\item[\algorithmicinput]}
\algnewcommand\algorithmicoutput{\textbf{Output:}}
\algnewcommand\Output{\item[\algorithmicoutput]}
\begin{algorithm}
\caption{Naive bootstrap of $\widehat{\mathrm{ROC}}_n(\alpha)$}\label{alg:naive_ROC}
\begin{algorithmic}
\Input $K \geq 1$, images $(X_1^{(1)}, \dotsc, X_{n_K}^{(K)})$, encoder $f$
\Require $\alpha \in (0,1)$, $B \geq 1$
\Output $B$ naive bootstrap versions $\widehat{\mathrm{ROC}}_n^*(\alpha) = \big(\widehat{\mathrm{ROC}}_{n, (b)}^*(\alpha)\big)_{1 \leq b \leq B}$ of the empirical ROC
\State $\widehat{\mathrm{ROC}}_{n}^*(\alpha) \gets \emptyset$
\For{$b\gets 1, \dotsc, B$}
\State $X_{(b)} \gets \emptyset$
\For{$k\gets 1, \dotsc, K$}
\State $X_{(b)}^{(k)} \gets$ sample with replacement $n_k$ images among $(X_1^{(k)}, \dotsc, X_{n_k}^{(k)})$
\State $X_{(b)} \gets X_{(b)} \cup X_{(b)}^{(k)}$
\EndFor
\State $\widehat{\mathrm{ROC}}_{n, (b)}^*(\alpha) \gets \widehat{\frr}^*_{n}\circ (\widehat{\far}^*_{n})^{-1}(\alpha)$ for bootstrap sample $X_{(b)}$
\State $\widehat{\mathrm{ROC}}_{n}^*(\alpha) \gets \widehat{\mathrm{ROC}}_{n}^*(\alpha) \cup \widehat{\mathrm{ROC}}_{n, (b)}^*(\alpha)$
\EndFor
\State \Return $\widehat{\mathrm{ROC}}_{n}^*(\alpha)$ 
\end{algorithmic}
\end{algorithm}


\paragraph{Naive bootstrap of fairness metrics.} Let us take the fairness metric $\mathrm{FRR}_{\mathrm{min}}^\mathrm{max}(\alpha)$ as an example. It is defined in section~\ref{subsec:roc} as: 
\[ \mathrm{FRR}_{\mathrm{min}}^\mathrm{max}(t) =  \frac{\max_{a \in \mathcal{A}} \mathrm{FRR}_a(t)}{\min_{a \in \mathcal{A}} \mathrm{FRR}_a(t)} \quad \text{with } t \text{ such that } \mathrm{FAR}(t)=\alpha.\]
Its empirical version can be defined (see \ref{app:empirical_fairness}), with the notations $\widehat{\frr}_{n,a}(t)$ and $\widehat{\far}_n(t)$ (defined in Equations~\ref{eq:emp_cdfG_a} and \ref{eq:emp_cdfH}), as:
\[ \widehat{\mathrm{FRR}}_{\mathrm{min,n}}^\mathrm{max}(\alpha) =  \frac{\max_{a \in \mathcal{A}} \widehat{\frr}_{n,a} \circ (\widehat{\far}_n)^{-1}(\alpha)}{\min_{a \in \mathcal{A}} \widehat{\frr}_{n,a}\circ (\widehat{\far}_n)^{-1}(\alpha)}.\]

To define the naive bootstrap of the quantity $\mathrm{FRR}_{\mathrm{min}}^\mathrm{max}(\alpha)$, it is necessary to first define the naive bootstrap of the quantity $\widehat{\frr}_{n,a}(t)$ (and $\widehat{\far}_{n,a}(t)$ if we consider FAR fairness metrics). The procedure is completely similar to the bootstrap of $\widehat{\frr}_n(t)$ and $\widehat{\far}_n(t)$, presented in section~\ref{subsec:bootstrap}.

The bootstrap paradigm suggests to recompute the quantities $\widehat{\frr}_{n,a}(t)$ and $\widehat{\far}_{n,a}(t)$ from independent sequences of i.i.d. variables $X^{(k)*}_{1},\; \ldots,\; X^{(k)*}_{n_k}$ with distribution 
\begin{equation}
\hat{F}_k=\frac{1}{n_k}\sum_{i=1}^{n_k}\delta_{X^{(k)}_i},
\end{equation}
conditioned upon the original dataset $\mathcal{D}=\{X^{(k)}_{1},\; \ldots,\; X^{(k)}_{n_k}:\; k=1,\; \ldots,\; K\}$. In practice of course, the resampling scheme would be applied $B\geq 1$ times in order to compute Monte-Carlo approximations of the distributions of 
\begin{eqnarray}\label{eq:boot_cdfG_a}
\widehat{\frr}^*_{n,a}(t)&=&\frac{1}{K_a}\sum_{\substack{k=1 \\ a_k = a}}^K\frac{2}{n_k(n_k-1)} \sum_{1\leq i<j\leq n_k}\mathbb{I}\{ s(X^{(k)*}_{i},X^{(k)*}_{j})\leq t  \}, \\
\label{eq:boot_cdfH_a}
\widehat{\far}^*_{n,a}(t)&=&\frac{2}{K_a(K_a-1)}\sum_{\substack{k< l \\ a_k = a_l = a}}\frac{1}{n_kn_l} \sum_{i=1}^{n_k}\sum_{j=1}^{n_l}\mathbb{I}\{ s(X^{(k)*}_{i},X^{(l)*}_{j})> t  \}.
\end{eqnarray}

Those boostrap versions of $\widehat{\frr}_{n,a}(t)$ and $\widehat{\far}_{n,a}(t)$ allow to define a bootstrap version of each fairness metric. In our running example, the fairness metric $\mathrm{FRR}_{\mathrm{min}}^\mathrm{max}(\alpha)$, one may compute its bootstrap version
\begin{equation}\label{boot_fairness}
\widehat{\mathrm{FRR}}_{\mathrm{min},n}^\mathrm{max *}(\alpha) =  \frac{\max_{a \in \mathcal{A}} \widehat{\frr}_{n,a}^* \circ (\widehat{\far}_n^*)^{-1}(\alpha)}{\min_{a \in \mathcal{A}} \widehat{\frr}_{n,a}^*\circ (\widehat{\far}_n^*)^{-1}(\alpha)}.
\end{equation}

The naive bootstrap for the fairness metric $\mathrm{FRR}_{\mathrm{min}}^\mathrm{max}(\alpha)$ follows exactly the same algorithm as Algorithm~\ref{alg:naive_ROC}, except that the variable $\widehat{\mathrm{ROC}}_{n}^*(\alpha)$ is now the bootstrap fairness metric $\widehat{\mathrm{FRR}}_{\mathrm{min},n}^\mathrm{max *}(\alpha)$ and the line $\widehat{\mathrm{ROC}}_{n, (b)}^*(\alpha) \gets \widehat{\frr}^*_{n}\circ (\widehat{\far}^*_{n})^{-1}(\alpha)$ becomes 
\[\widehat{\mathrm{FRR}}_{\mathrm{min}, n, (b)}^\mathrm{max *}(\alpha) \gets \frac{\max_{a \in \mathcal{A}} \widehat{\frr}_{n,a}^* \circ (\widehat{\far}_n^*)^{-1}(\alpha)}{\min_{a \in \mathcal{A}} \widehat{\frr}_{n,a}^*\circ (\widehat{\far}_n^*)^{-1}(\alpha)}.\]



\subsection{Pseudo-code for the computation of a confidence interval}

Algorithm~\ref{alg:CI} provides a pseudo-code for the computation of a confidence interval for $\widehat{\mathrm{ROC}}_n(\alpha)$ at level $1-\alpha_{CI}$. It requires the output of Algorithm~\ref{alg:naive_ROC} which gives $B$ naive bootstrap versions $\big(\widehat{\mathrm{ROC}}_{n, (b)}^*(\alpha)\big)_{1 \leq b \leq B}$ of the empirical ROC.
As explained in section~\ref{subsec:bootstrap}, the recentered bootstrap for $\widehat{\mathrm{ROC}}_n(\alpha)$ consists in taking each naive boostrap version $\widehat{\mathrm{ROC}}_{n, (b)}^*(\alpha)$, computing its distance to the V-statistic counterpart $\widetilde{\mathrm{ROC}}_n(\alpha) = \widetilde{\frr}_{n}\circ (\widehat{\far}_{n})^{-1}(\alpha)$, and then shifting it by $\widehat{\mathrm{ROC}}_n(\alpha)$. This is exactly what Algorithm~\ref{alg:CI} does, but the shift is done after computation of the quantiles (for the confidence interval). For each naive bootstrap version $\widehat{\mathrm{ROC}}_{n, (b)}^*(\alpha)$, we compute the gap with respect to the V-statistic $\widetilde{\mathrm{ROC}}_n(\alpha)$, and accumulate, for all bootstrap steps $b$, those distances into the variable gap. Then, a confidence interval at level $1-\alpha_{CI}$ is computed, giving one lower bound $l$ and one upper bound $u$, defining the confidence interval. Eventually, this confidence interval is shifted by $\widehat{\mathrm{ROC}}_n(\alpha)$ as mentioned earlier. 

\begin{algorithm}\label{alg:ci}
\caption{Confidence interval for $\widehat{\mathrm{ROC}}_n(\alpha)$}\label{alg:CI}
\begin{algorithmic}
\Input $K \geq 1$, images $(X_1^{(1)}, \dotsc, X_{n_K}^{(K)})$, encoder $f$
\Require {$\alpha \in (0,1)$, $B \geq 1$, $\alpha_{CI} \in (0,1)$ \\
\quad \quad \quad $\widehat{\mathrm{ROC}}_{n}^*(\alpha) = \big(\widehat{\mathrm{ROC}}_{n, (b)}^*(\alpha)\big)_{1 \leq b \leq B}$ from \autoref{alg:naive_ROC}}
\Output $l \text{ and } u$, bounds for the confidence interval of $\widehat{\mathrm{ROC}}_n(\alpha)$ at level $1-\alpha_{CI}$
\State $\widetilde{\mathrm{ROC}}_n(\alpha) \gets \widetilde{\frr}_{n}\circ (\widehat{\far}_{n})^{-1}(\alpha)$ for original data $(X_1^{(1)}, \dotsc, X_{n_K}^{(K)})$
\State $\text{gap} \gets \emptyset$
\For{$b\gets 1, \dotsc, B$}
\State $\text{gap}_{(b)} \gets \widehat{\mathrm{ROC}}_{n, (b)}^*(\alpha) - \widetilde{\mathrm{ROC}}_n(\alpha)$
\State $\text{gap} \gets \text{gap} \cup \text{gap}_{(b)}$
\EndFor
\State $l \gets$ $\displaystyle \frac{\alpha_{CI}}{2}$-th quantile of gap
\State $u \gets$ $(1-\displaystyle \frac{\alpha_{CI}}{2})$-th quantile of gap

\State $\widehat{\mathrm{ROC}}_n(\alpha) \gets \widehat{\frr}_{n}\circ (\widehat{\far}_{n})^{-1}(\alpha)$ for original data $(X_1^{(1)}, \dotsc, X_{n_K}^{(K)})$

\State $l \gets \widehat{\mathrm{ROC}}_n(\alpha) + l$
\State $u \gets \widehat{\mathrm{ROC}}_n(\alpha) + u$
\State \Return $l, u$ 
\end{algorithmic}
\end{algorithm}


\paragraph{Confidence interval for fairness metrics.}  Let us take the fairness metric $\mathrm{FRR}_{\mathrm{min}}^\mathrm{max}(\alpha)$ as an example. It is defined in section~\ref{subsec:roc} as: 
\[ \mathrm{FRR}_{\mathrm{min}}^\mathrm{max}(t) =  \frac{\max_{a \in \mathcal{A}} \mathrm{FRR}_a(t)}{\min_{a \in \mathcal{A}} \mathrm{FRR}_a(t)} \quad \text{with } t \text{ such that } \mathrm{FAR}(t)=\alpha.\]
Its empirical version can be defined (see \ref{app:empirical_fairness}), with the notations $\widehat{\frr}_{n,a}(t)$ and $\widehat{\far}_n(t)$ (defined in Equations~\ref{eq:emp_cdfG_a} and \ref{eq:emp_cdfH}), as:
\[ \widehat{\mathrm{FRR}}_{\mathrm{min,n}}^\mathrm{max}(\alpha) =  \frac{\max_{a \in \mathcal{A}} \widehat{\frr}_{n,a} \circ (\widehat{\far}_n)^{-1}(\alpha)}{\min_{a \in \mathcal{A}} \widehat{\frr}_{n,a}\circ (\widehat{\far}_n)^{-1}(\alpha)}.\]

To achieve the computation of a confidence interval, the method to do it for the ROC curve, presented above, uses the V-statistic counterpart. In the same way as for $\widehat{\frr}_n(t)$ (see section~\ref{subsec:bootstrap}), we can define the V-statistic counterpart for $\widehat{\frr}_{n,a}(t)$ as
\begin{equation}\label{eq:Vstat_G_a}    
\widetilde{\frr}_{n,a}(t):=\frac{1}{K_a}\sum_{\substack{k=1 \\ a_k = a}}^K\frac{1}{n^2_k} \sum_{1\leq i,j\leq n_k}\mathbb{I}\{ s(X^{(k)}_{i},X^{(k)}_{j})\leq t  \}.
\end{equation}

As for the ROC curve, this allows us to define the V-statistic counterpart of the fairness metric $\mathrm{FRR}_{\mathrm{min}}^\mathrm{max}(\alpha)$ as
\begin{equation}\label{eq:Vstat_fairness}
\widetilde{\mathrm{FRR}}_{\mathrm{min},n}^\mathrm{max}(\alpha) =  \frac{\max_{a \in \mathcal{A}} \widetilde{\frr}_{n,a} \circ (\widehat{\far}_n)^{-1}(\alpha)}{\min_{a \in \mathcal{A}} \widetilde{\frr}_{n,a}\circ (\widehat{\far}_n)^{-1}(\alpha)}.    
\end{equation}

The computation of a confidence interval for the quantity $\mathrm{FRR}_{\mathrm{min}}^\mathrm{max}(\alpha)$ follows exactly the same algorithm as Algorithm~\ref{alg:CI}. We require the output $\widehat{\mathrm{FRR}}_{\mathrm{min}, n, (b)}^\mathrm{max *}(\alpha)$, instead of $\widehat{\mathrm{ROC}}_{n, (b)}^*(\alpha)$, for $b = 1, \dotsc, B$, of Algorithm~\ref{alg:naive_ROC} applied to $\mathrm{FRR}_{\mathrm{min}}^\mathrm{max}(\alpha)$. For the V-statistic, the line $\widetilde{\mathrm{ROC}}_n(\alpha) \gets \widetilde{\frr}_{n}\circ (\widehat{\far}_{n})^{-1}(\alpha)$ becomes 
\[ \widetilde{\mathrm{FRR}}_{\mathrm{min},n}^\mathrm{max}(\alpha) \gets \frac{\max_{a \in \mathcal{A}} \widetilde{\frr}_{n,a} \circ (\widehat{\far}_n)^{-1}(\alpha)}{\min_{a \in \mathcal{A}} \widetilde{\frr}_{n,a}\circ (\widehat{\far}_n)^{-1}(\alpha)}, \] 
hence the gap now measures the distance between $\widehat{\mathrm{FRR}}_{\mathrm{min}, n, (b)}^\mathrm{max *}(\alpha)$ and $\widetilde{\mathrm{FRR}}_{\mathrm{min},n}^\mathrm{max}(\alpha)$. For the empirical part, the line $\widehat{\mathrm{ROC}}_n(\alpha) \gets \widehat{\frr}_{n}\circ (\widehat{\far}_{n})^{-1}(\alpha)$ becomes 
\[ \widehat{\mathrm{FRR}}_{\mathrm{min},n}^\mathrm{max}(\alpha) \gets \frac{\max_{a \in \mathcal{A}} \widehat{\frr}_{n,a} \circ (\widehat{\far}_n)^{-1}(\alpha)}{\min_{a \in \mathcal{A}} \widehat{\frr}_{n,a}\circ (\widehat{\far}_n)^{-1}(\alpha)},\]
as well as for the ending lines shifting bounds of the confidence interval. 

\subsection{Pseudo-code for the computation of the uncertainty} 

Algorithm~\ref{alg:uncertainty} provides a pseudo-code for the computation of the normalized uncertainty (Eq.~\ref{eq:normalized_uncertainty}) for the $\roc$ curve. It requires the output of Algorithm~\ref{alg:naive_ROC} which gives $B$ naive bootstrap versions $\big(\widehat{\mathrm{ROC}}_{n, (b)}^*(\alpha)\big)_{1 \leq b \leq B}$ of the empirical ROC. As explained in section~\ref{subsec:bootstrap}, the standard deviation of each of those $B$ boostrapped values (minus the V-statistic counterpart) is computed and then divided by $\widehat{\mathrm{ROC}}_{n}(\alpha)$. The algorithm is quite similar to the one used to compute confidence intervals (Algorithm~\ref{alg:CI}). 

\begin{algorithm}
\caption{Uncertainty of $\widehat{\mathrm{ROC}}_n(\alpha)$}\label{alg:uncertainty}
\begin{algorithmic}
\Input $K \geq 1$, images $(X_1^{(1)}, \dotsc, X_{n_K}^{(K)})$, encoder $f$
\Require {$\alpha \in (0,1)$, $B \geq 1$\\
\quad \quad \quad $\widehat{\mathrm{ROC}}_{n}^*(\alpha) = \big(\widehat{\mathrm{ROC}}_{n, (b)}^*(\alpha)\big)_{1 \leq b \leq B}$ from \autoref{alg:naive_ROC}}
\Output $U$, normalized uncertainty of $\widehat{\mathrm{ROC}}_n(\alpha)$
\State $\widetilde{\mathrm{ROC}}_n(\alpha) \gets \widetilde{\frr}_{n}\circ (\widehat{\far}_{n})^{-1}(\alpha)$ for original data $(X_1^{(1)}, \dotsc, X_{n_K}^{(K)})$
\State $\text{gap} \gets \emptyset$
\For{$b\gets 1, \dotsc, B$}
\State $\text{gap}_{(b)} \gets \widehat{\mathrm{ROC}}_{n, (b)}^*(\alpha) - \widetilde{\mathrm{ROC}}_n(\alpha)$
\State $\text{gap} \gets \text{gap} \cup \text{gap}_{(b)}$
\EndFor
\State $U \gets$ standard deviation of gap

\State $\widehat{\mathrm{ROC}}_n(\alpha) \gets \widehat{\frr}_{n}\circ (\widehat{\far}_{n})^{-1}(\alpha)$ for original data $(X_1^{(1)}, \dotsc, X_{n_K}^{(K)})$

\State $U \gets U / \widehat{\mathrm{ROC}}_n(\alpha)$
\State \Return $U$ 
\end{algorithmic}
\end{algorithm}

\paragraph{Uncertainty of fairness metrics.} The extension to those quantities is completely similar to the extension detailed for Algorithm~\ref{alg:CI}. The computation of the normalized uncertainty for all fairness metrics is provided in Figure~\ref{fig:std_fairness} and allows to compare the uncertainty of each fairness metric to exhibit their relative robustness.

\section{Technical Details}

\subsection{A Note on the Pseudo-Inverse of the $\far$ quantity}\label{app:inverse_far}

In \ref{subsec:roc}, we introduced the $\far$ metric, defined as:
\[\far(t)=\mathbb{P}\{ s(X,X') > t  \mid Z=-1 \},\]
and the $\roc$ curve as $\roc \colon \alpha \in (0,1)\mapsto \frr \circ \far^{-1}(\alpha)$.

The pseudo-inverse of any cumulative distribution function (cdf) $\kappa(t)$ on $\mathbb{R}$ is defined as 
\[\kappa^{-1}(\alpha)=\inf\{t\in \mathbb{R}:\; \kappa(t)\geq \alpha  \}, \quad \text{for } \alpha \in (0,1).\]
Note that the quantity $\far(t)$ is not a cdf, so that its pseudo-inverse is not well defined. However, the opposite of $\far(t)$, the True Rejection Rate ($\trr$), is a proper cdf:
\[\trr(t)=1-\far(t)=\mathbb{P}\{ s(X,X') \leq t  \mid Z=-1 \}.\]
As such, the pseudo-inverse $\trr^{-1}(\alpha)$ is well defined for the $\trr$ quantity. This allows to define the pseudo-inverse for $\far$. Indeed, for any $\alpha \in (0,1)$ satisfying $\far(t)=\alpha$, one would get the following $\trr$
\[
\trr(t) = 1-\alpha.
\]
The threshold $t$ of interest is found using the pseudo-inverse of $\trr$ (see \cite{Hsieh1}):
\[
\far^{-1}(\alpha) \colon= \trr^{-1}(1-\alpha) = (1-\far)^{-1}(1-\alpha).
\]
This definition is extended to other quantities within the paper, all being the opposite of one cdf.

\subsection{Proof of Proposition \ref{prop:limit} - Consistency of the Empirical Similarity $\roc$ Curve}\label{app:consistency_proof}

In the multi-sample asymptotic framework \ref{subsec:framework}, by virtue of the $U$-statistic's version of the Strong Law of Large Numbers (see \cite{Ser80} or \cite{Lee1990}), we almost-surely have:
\begin{equation*}
\widehat{\far}_n(t)\to \far(t) \text{ and } \widehat{\frr}_n(t)\to \frr(t) \text{ as } n\to +\infty,
\end{equation*}
for all $t\in \mathbb{R}$. Applying next the argument of Lemma 21.2 in \cite{vdV98}, we deduce that, for all $\alpha\in (0,1)$, we have with probability one:
\[(\widehat{\far}_n)^{-1}(\alpha)\to \far^{-1}(\alpha) \text{ as } n\to +\infty.\]
We thus obtain the pointwise consistency of the empirical similarity $\roc$ curve, the uniform version being immediately obtained by a classic Dini's argument, given the monotone nature of (empirical) $\roc$ curves.

Notice incidentally that a bound of order $O_{\mathbb{P}}(1/\sqrt{n})$ could be established by means of the same linearization techniques (\textit{i.e.} Hoeffding decomposition) as those used in \cite{pmlr-v80-vogel18a}. 

\subsection{Definitions of $\widehat{\far}_{n,a}(t)$, $\widehat{\frr}_{n,a}(t)$ and empirical fairness metrics}\label{app:empirical_fairness}

 Without specifying any particular subgroup $a$ (\textit{i.e.} considering the global poulation), we have already explained in section~\ref{subsec:consistency} that $\widehat{\far}_n(t)$ and $\widehat{\frr}_n(t)$ (defined in Equations~\ref{eq:emp_cdfH} and \ref{eq:emp_cdfG}) are natural empirical versions of $\far(t)$ and $\frr(t)$. We extend here their definitions to one specific subgroup $a \in \mathcal{A}$. For this purpose, we first define $a_k \in \mathcal{A}$ the (sensitive) attribute label associated with identity $k$, that is the subgroup to which all images of identity $k$ belong to. We also define $K_a$ the number of identities which belong to subgroup $a \in \mathcal{A}$, within the evaluation dataset. For any $a \in \mathcal{A}$, we then define the natural empirical versions of $\far_a(t)$ and $\frr_a(t)$:

\begin{eqnarray}\label{eq:emp_cdfG_a}
\widehat{\frr}_{n,a}(t)&=&\frac{1}{K_a}\sum_{\substack{k=1 \\ a_k = a}}^K\frac{2}{n_k(n_k-1)} \sum_{1\leq i<j\leq n_k}\mathbb{I}\{ s(X^{(k)}_{i},X^{(k)}_{j})\leq t  \},\\
\label{eq:emp_cdfH_a}
\widehat{\far}_{n,a}(t)&=&\frac{2}{K_a(K_a-1)}\sum_{\substack{k< l \\ a_k = a_l =a}}\frac{1}{n_kn_l} \sum_{i=1}^{n_k}\sum_{j=1}^{n_l}\mathbb{I}\{ s(X^{(k)}_{i},X^{(l)}_{j}) > t  \}.
\end{eqnarray}

Note that they are still U-statistics, as for $\widehat{\far}_n(t)$ and $\widehat{\frr}_n(t)$.

The empirical versions of the fairness metrics listed in section~\ref{app:fairness_metrics} can then be expressed in terms of $\widehat{\far}_{n,a}(t), \widehat{\frr}_{n,a}(t)$ and $\widehat{\far}_n(t)$. Let us take the fairness metric $\mathrm{FRR}_{\mathrm{min}}^\mathrm{max}(\alpha)$ as an example. It is defined in section~\ref{app:fairness_metrics} as: 
\[ \mathrm{FRR}_{\mathrm{min}}^\mathrm{max}(\alpha) =  \frac{\max_{a \in \mathcal{A}} \mathrm{FRR}_a \circ \far^{-1}(\alpha)}{\min_{a \in \mathcal{A}} \mathrm{FRR}_a \circ \far^{-1}(\alpha)}.\]

Its empirical version is simply defined as:

\[ \widehat{\mathrm{FRR}}_{\mathrm{min}, n}^\mathrm{max}(\alpha) =  \frac{\max_{a \in \mathcal{A}} \widehat{\frr}_{n,a} \circ (\widehat{\far}_n)^{-1}(\alpha)}{\min_{a \in \mathcal{A}} \widehat{\frr}_{n,a}\circ (\widehat{\far}_n)^{-1}(\alpha)}.\]

\subsection{Assumptions for next results}\label{app:assumptions}

The following results hold under the classic (mild) assumptions below.

As in \ref{app:inverse_far}, we still define $\trr$ as the cdf associated to the $\far$ quantity.

\begin{assumption}\label{hyp1} The univariate distributions $\trr$ and $\frr$ have densities $h$ and $g$ respectively and the slope of the $\roc$ curve is bounded: $\sup_{\alpha \in [0,1]}\{g(\trr^{-1}(\alpha))/h(\trr^{-1}(\alpha))\}<\infty$.
\end{assumption}
\begin{assumption}\label{hyp2}
    The cdf $\trr$ is twice differentiable on $[0,1]$ and $\forall \alpha \in [0,1],\;\; h(\alpha)>0$
and there exists $\gamma>0$ such that $\sup_{\alpha \in [0,1]}\{ \alpha(1-\alpha)\cdot d\log(h\circ \trr^{-1}(\alpha))/d\alpha
\}\leq\gamma<\infty$.
\end{assumption}

\subsection{Asymptotic validity of the naive and recentered bootstraps}\label{app:validity_bootstrap}

The result below also states the asymptotic validity of the naive bootstrap, described in \ref{subsec:bootstrap}.

\begin{theorem}\label{thm:naive_boot_val} 
Suppose that Assumptions \ref{hyp1}-\ref{hyp2} are satisfied. Then, for all $\alpha \in (0,1)$, we almost-surely have:
\begin{equation}\label{eq:limit_naive}
\sup_{v\in \mathbb{R}}\left\vert \mathbb{P}^*\left\{  \sqrt{n}\vert\widehat{\roc}_n^*(\alpha)- \widehat{\roc}_n(\alpha)\vert\leq v \mid \mathcal{D} \right\} - \mathbb{P}\left\{\sqrt{n}\vert \widehat{\roc}_n(\alpha) - \roc(\alpha)\vert \leq v \right\} \right\vert \rightarrow 0,
\end{equation}
as $n\rightarrow +\infty$.
\end{theorem}
The proof is the same as that of Theorem~\ref{thm:boot_val}.

The result below states the asymptotic validity of the recentered bootstrap for the $\roc$ curve, described in \ref{subsec:bootstrap}. As explained within the proof, the result holds when replacing the $\roc$ curve by fairness metrics. Indeed, one can define the bootstrap fairness metrics, as well as the $V$-statistic version of those metrics, in the same way than for the $\roc$ curve.

\begin{theorem}\label{thm:boot_val}
Suppose that Assumptions \ref{hyp1}-\ref{hyp2} are satisfied. Then, for all $\alpha \in (0,1)$, we almost-surely have:
\begin{equation}\label{eq:limit}
\sup_{v\in \mathbb{R}}\left\vert \mathbb{P}^*\left\{  \sqrt{n}\vert \widehat{\roc}_n^*(\alpha)- \widetilde{\roc}_n(\alpha) \vert \leq v \mid \mathcal{D} \right\} - \mathbb{P}\left\{\sqrt{n}\vert \widehat{\roc}_n(\alpha) - \roc(\alpha) \vert \leq v \right\} \right\vert \rightarrow 0,
\end{equation}
as $n\rightarrow +\infty$.
\end{theorem}
\begin{proof}
The asymptotic validity of the naive bootstrap procedure for non-degenerate (generalized, multivariate) $U$-statistics is proved in \cite{10.2307/2240410} (see section 3 therein, refer also to Theorem 3 in \cite{janssen1997bootstrapping} and to Theorem 3.3 in \cite{ShaoTu}). The proof is based on their asymptotic Gaussianity as well as that of the related $V$-statistics in the asymptotic framework \ref{subsec:framework} (refer to \textit{e.g.} \cite{Lee1990}), which holds true under the assumptions stipulated, combined with a coupling argument. Hence, with probability one, the absolute deviation between the bootstrap approximation
\begin{equation*}\label{eq:boot_val_cdf}
 \mathbb{P}^*\left\{  \sqrt{n}\left(\widehat{\far}_n^*(t_1)- \widehat{\far}_n(t_1)\right)\leq u,\;  \sqrt{n}\left(\widehat{\frr}_n^*(t_2)- \widetilde{\frr}_n(t_2)\right)\leq v \mid \mathcal{D} \right\} 
 \end{equation*}
 and the root distribution
  \begin{equation*}\mathbb{P}\left\{\sqrt{n}\left( \widehat{\far}_n(t_1) - \far(t_1) \right) \leq u,\; \sqrt{n}\left( \widehat{\frr}_n(t_2) - \frr(t_2) \right) \leq v \right\}
\end{equation*}
converges to $0$, uniformly in $(u,v)\in \mathbb{R}^2$, as $n\to \infty$. As noticed in \cite{janssen1997bootstrapping}, the bootstrapped FRR statistic $\widehat{\frr}_n^*(t_2)$ can be recentered either by the $V$-statistic version $\widetilde{\frr}_n(t_2)$ or else by the original $U$-statistic $\widehat{\frr}_n(t_2)$ due to the non-degeneracy property.

In addition, under the hypotheses stipulated, the asymptotic normality of the bivariate random vectors 
$$
\sqrt{n}\left( (\widehat{\far}_n)^{-1}(\alpha) - \far^{-1}(\alpha) ,\;  \widehat{\frr}_n(t) - \frr(t) \right)
$$
can be classically deduced from that of the random vectors
$$
\sqrt{n}\left( \widehat{\far}_n(t_1) - \far(t_1) ,\;  \widehat{\frr}_n(t_2) - \frr(t_2) \right)
$$
by means of classic (linearization) arguments for empirical quantiles (see \textit{e.g.} Chapter 21 in \cite{vdV98}). A Central Limit Theorem for $\sqrt{n}(\widehat{\roc}_n(\alpha)-\roc(\alpha))$ can be easily deduced and exactly the same argument as the one developed in \cite{10.2307/2240410} can then be used to prove (\ref{eq:limit}).

In a similar fashion, the asymptotic validity of the naive/recentered bootstrap applied to Hadamard differentiable functionals of the random vector $(\widehat{\far}_{n,a}(t), \widehat{\frr}_{n,a}(t'))_{a\in \mathcal{A}}$, such as the fairness metrics considered here, results from the asymptotic normality property combined with Theorem 23.9 in \cite{vdV98}.
\end{proof}

\subsection{Consistency of the coverage provided by the confidence intervals}\label{app:consistency_ci} 

It results from Theorem \ref{thm:boot_val}.

\begin{corollary}\label{corro:consistency_ci}
Let $\alpha\in (0,1)$ and $\alpha_{CI}\in (0,1)$. Under the assumptions of Theorem \ref{thm:boot_val}, we have:
\[
\mathbb{P}\{ l_{\alpha_{CI}}^{(n,B)}(\alpha) \leq \roc(\alpha) \leq u_{\alpha_{CI}}^{(n,B)}(\alpha)\} \rightarrow 1-\alpha_{CI},
\]
as $n$ and $B$ both tend to $+\infty$.
\end{corollary}

 \begin{proof}
     The consistency of the probability coverages of the Monte Carlo confidence intervals based on $B\geq 1$ bootstrap replicates and described in subsection \ref{subsec:bootstrap} immediately results from Theorem \ref{thm:boot_val}, combined with the Strong Law of Large Numbers.
 \end{proof}

 \subsection{Bootstrap and normalized uncertainty of fairness metrics}\label{app:fairness_bootstrap}

We have illustrated all results with the use case of the $\roc$ curve but those results hold for the fairness metrics listed in \ref{app:fairness_metrics}. One only has to replace the $\roc$-related quantities by their fairness counterparts:
\begin{itemize}
    \item $\roc(\alpha)$: the true fairness metrics have been defined in \ref{app:fairness_metrics}.
    \item $\widehat{\roc}_n(\alpha)$: the empirical fairness metrics are defined in \ref{app:empirical_fairness}. Their definition is very similar to $\widehat{\roc}_n(\alpha)$ in the sense that one has to replace the $\far$, $\frr$ quantities by their empirical versions (\textit{plug-in}).
    \item $\widehat{\roc}_n^*(\alpha)$: the bootstrap fairness metrics are nothing but the empirical fairness metrics, computed with a bootstrap sample, instead of the original dataset (exactly like the $\roc$ curve).
    \item $\widetilde{\roc}_n(\alpha)$: the $V$-statistic version of fairness metrics is simply obtained by replacing $\far_a(t)$ with its empirical version $\widehat{\far}_{n,a}(t)$ (see \ref{app:empirical_fairness}) and $\frr_a(t)$ with its $V$-statistic version, exactly as for the $\roc$ curve.
\end{itemize}

{\bf Bootstrap and confidence intervals.} As explained within the proof of Theorem~\ref{thm:boot_val}, the naive/recentered bootstrap validity holds for fairness metrics when replacing the $\roc$-related quantities with the fairness quantities listed above. The consistence of the confidence intervals (Corollary~\ref{corro:consistency_ci}) directly results from Theorem~\ref{thm:boot_val}.

{\bf Normalized uncertainty.} In \ref{subsec:bootstrap}, we defined a scalar uncertainty measure for the $\roc$ curve as:
\begin{equation}
U[\widehat{\roc}_n(\alpha)]  = \frac{\sqrt{\mathrm{Var}[\hat{\epsilon}_n^{(2)}(\alpha)\mid \mathcal{D}]}}{\widehat{\roc}_n(\alpha)}, 
\end{equation}
with $\hat{\epsilon}_n^{(2)}(\alpha) =\widehat{\roc}_n^*(\alpha)-\widetilde{\roc}_n(\alpha)$. Replacing $\widehat{\roc}_n^*(\alpha)$, $\widetilde{\roc}_n(\alpha)$ by their fairness counterparts (see above) allows to get a fairness version of $\hat{\epsilon}_n^{(2)}(\alpha)$. Then, to get the normalized uncertainty of a fairness measure, one only has to compute the (root of) variance of this fairness version of $\hat{\epsilon}_n^{(2)}(\alpha)$, and then to normalize by the corresponding empirical fairness measure. 
As explained within the proof of Theorem~\ref{thm:boot_val}, the naive/recentered bootstrap validity holds for fairness metrics when replacing the $\roc$-related quantities with the fairness quantities listed above. Thus, the normalized uncertainty is as pertinent to fairness metrics than it is to the $\roc$ curve.

Note that we give the methodology (bootstrap, confidence intervals, normalized uncertainty) step by step, for one fairness metric, in the pseudo-code \ref{app:code}.

\end{document}